\documentclass[10pt,journal,compsoc]{IEEEtran}
\usepackage{mathptmx} % This is Times font

\usepackage{fancyhdr}
\usepackage[normalem]{ulem}
\usepackage[hyphens]{url}
\usepackage[sort,nocompress]{cite}
\usepackage[final]{microtype}
\usepackage[keeplastbox]{flushend}
\usepackage[bookmarks=true,breaklinks=true,letterpaper=true,colorlinks,citecolor=black,linkcolor=black,urlcolor=black]{hyperref}
\usepackage{tikz}
\usepackage{tablefootnote}
\usepackage{amsmath}
\usepackage{color}
\usepackage{amssymb}
\usepackage{comment}

\usepackage{multirow}
\usepackage{array}
\usepackage{threeparttable}

\ifCLASSOPTIONcompsoc
  \usepackage[nocompress]{cite}
\else
  \usepackage{cite}
\fi
\ifCLASSINFOpdf
\else
\fi
\begin{document}
\title{Real-Time Robust Video Object Detection System Against Physical-World \\ Adversarial Attacks} %\Large \bf

% ~\IEEEmembership{Member,~IEEE,}
\author{Husheng~Han,Xing~Hu~\IEEEmembership{IEEE,}
        Kaidi~Xu,Pucheng~Dang,Ying~Wang,
        Yongwei~Zhao,Zidong~Du~\IEEEmembership{IEEE,} Qi~Guo~\IEEEmembership{IEEE,} Yanzhi~Yang,Tianshi~Chen
        % and~Jane~Doe,~\IEEEmembership{Life~Fellow,~IEEE}% <-this % stops a space
\thanks{The preliminary version is published in NeurIPS 2021.}
\IEEEcompsocitemizethanks{
\IEEEcompsocthanksitem Husheng Han, Pucheng Dang and Yongwei Zhao are with the SKL of Processors, Institute of Computing Technology, CAS, Beijing 100190, China, the University of Chinese Academy of Sciences, Beijing, 100049, China, and also with Cambricon Technologies, Beijing, China. \protect\\ E-mail:\{hanhusheng20z,dangpucheng20g,zhaoyongwei\}@ict.ac.cn
\IEEEcompsocthanksitem Xing Hu and Qi Guo are with the SKL of Processors, Institute of Computing Technology, CAS, Beijing, 100190, China. \protect\\
E-mail:\{huxing,guoqi\}@ict.ac.cn
\IEEEcompsocthanksitem Kaidi Xu is with Department of Computer Science, College of Computing \& Informatics, Drexel University. Philadelphia, 19104,  USA.  \protect\\ E-mail:kx46@drexel.edu
\IEEEcompsocthanksitem Zidong Du is with the SKL of Processors, Institute of Computing Technology, CAS, Beijing, 100190, China, and also with Cambricon Technologies, Beijing, China. \protect 
E-mail:duzidong@ict.ac.cn
\IEEEcompsocthanksitem Tianshi Chen is with Cambricon Technologies, Beijing, China. \protect\\
E-mail:chentianshi@ict.ac.cn
\IEEEcompsocthanksitem Ying Wang is with SKL of Computer Architecture, Institute of Computing Technology, CAS, Beijing 100190, China. Email:wangying2009@ict.ac.cn
\IEEEcompsocthanksitem Yanzhi Yang is with  Department of Electrical and Computer Engineering, Northeastern University, Boston, 02115, USA  \protect\\ Email:yanz.wang@northeastern.edu
}% <-this % stops an unwanted space
% \thanks{Manuscript received April 19, 2005; revised August 26, 2015.}
\thanks{(Corresponding author: Xing Hu.)}
}


% \markboth{Journal of \LaTeX\ Class Files,~Vol.~14, No.~8, August~2015}%
\markboth{}%
{Shell \MakeLowercase{\textit{et al.}}: Bare Demo of IEEEtran.cls for Computer Society Journals}



\IEEEtitleabstractindextext{%
\begin{abstract}

DNN-based video object detection (VOD) powers autonomous driving and video surveillance industries with rising importance and promising opportunities. However, adversarial patch attack yields huge concern in live vision tasks because of its practicality, feasibility, and powerful attack effectiveness. This work proposes \textit{Themis}, a software/hardware system to defend against adversarial patches for real-time robust video object detection. We observe that adversarial patches exhibit extremely localized superficial feature importance in a small region with non-robust predictions, and thus propose the adversarial region detection algorithm for adversarial effect elimination.
\textit{Themis} also proposes a systematic design to efficiently support the algorithm by eliminating redundant computations and memory traffics. Experimental results show that the proposed methodology can effectively recover the system from the adversarial attack with negligible hardware overhead.
\end{abstract}

}

\maketitle

\IEEEdisplaynontitleabstractindextext

% \fi
\IEEEpeerreviewmaketitle

% \IEEEraisesectionheading{\section{Introduction}\label{sec:introduction}}

% --------------------------------------------------------------------------------

% \begin{document}
% \maketitle
% \thispagestyle{firstpage}
% \pagestyle{plain}

%%%%%% -- PAPER CONTENT STARTS-- %%%%%%%%

\section{Introduction}

%\Note{The video detection boost the autonomous driving and }
\IEEEPARstart{P}{owered} by deep neural network (DNN) techniques, video recognition achieves tremendous success and starts to boost existing industries, such as autonomous driving, surveillance systems, drones, and robots.
%Deep neural networks (DNNs) revolutionize computer-vision applications~\cite{NIPS2012:Hinton,VGG2014,resnet} and start to power existing industries.
For example, autonomous driving based on video recognition, whose market is predicted to leap to \$77 billion (25\% of the whole automotive market) by 2035 \cite{gmc_report}, 
%with \$77 billion projected in revenue by 2035, 
has attracted the attention of giants including Tesla and Waymo~\cite{tesla_news,waymo1}. %tesla_news2, audi_nvidia

%\Note{Threats in the video detection}
%\subNote{Threat model and bad consequences}
Despite the promising opportunities and rising importance of DNN-powered video recognition, the vulnerability of DNNs emerges as an important problem to video recognition tasks, especially in life-critical scenarios.
DNN techniques have shown to be vulnerable to adversarial attacks. For example, wearing the T-shirt with adversarial patch printing on it, which effectively fools DNN-based person detectors in physical environments even under diverse scenarios like people walking, sitting, and running~\cite{xu2020adversarial,wu2020making}. Such attacks are malicious in the surveillance and autonomous vehicle application scenarios, which evade the video detectors in the physical world and incur life-or-death problems. Therefore, robust video recognition that defends against such adversarial attacks and eliminates the adversarial effects is urgent and important.

%\subNote{adversarial patch attack is the most malicious one because of the scene-independent feature and robustness}
%\subNote{Conclusion: the urgeness to save video detection systems from such attack}

%\Note{Limitation of related work, focus on the image detector instead of video, summarize the challenges when extension to video cases}
%\subNote{scene-dependent attack(adv example) vs. scene-independent attack (adv patch)}
%\subNote{Cannot meet the real-time demands. }

% \begin{figure}[t!]
% \centerline{\includegraphics[width=0.5\textwidth]{fig/advT.pdf}}
% \vspace{-10pt}
% \caption{Real-world samples of adversarial attacks in video recognition~\cite{xu2020adversarial}: the person wearing the T-shirt printed with the adversarial patch is ignored by the detector.}
% \vspace{-5pt}
% \label{fig:advT}
% \end{figure}

For live vision scenarios, the defensive methodology should meet the following two requirements: 1) Effectively recover (more than detection only) the system from the adversarial attacks considering video recognition is usually adopted in real-time decision-making scenarios. 
%, even under a strong attack model that the adversary has the white-box knowledge of the defensive strategy (which is referred to as adaptive attacks in the following); 
2)  The proposed defensive methodology should introduce lightweight performance overhead to achieve the goal of real-time object detection. 
Effectively recovering the VOD system from adversarial attack in real-time is a highly challenging task. 
Existing pioneering studies for robust image classification fail to meet these two requirements: 
%In observing the vulnerability of deep learning techniques, some pioneering studies propose defensive methodologies~\cite{gan2020ptolemy,wang2020dnnguard}. However, they fail to meet the effectiveness and efficiency requirements: 
They either detect abnormal inputs only without recovery %Previous studies leverage activation paths or activation map for abnormal input detection
~\cite{xiang2021patchguard++,abusnaina2021latentNG,tian2021Spatial,wang2022manda},
% lee2018simple, feinman2017detecting, ma2018characterizing, wang2020dnnguard, yang2020mlloo,gan2020ptolemy,

or introduce too much overhead that cannot be born in real-time VOD systems

%2) Only work on very small image dataset (MNIST or Cifar) instead of industrialized-level input images~\cite{chiang2020certified,zhang2020clipped,levine2020randomized}.

%3) Introduces large overhead that hardly applied in real-time VOD systems
~\cite{mrd,xiang2022objectseeker}. MRD introduces extremely large overhead (costs about 1446s for one ImageNet-class image), which is not feasible in real-time scenario~\cite{mrd}. Patchguard++ relies on analyzing deep feature of every single image and cannot utilize the temporal and spatial redundancy of adjacent frames in video data~\cite{xiang2020patchguard}. 

%4) Rely on analyzing deep features of every single image, which can hardly utilize the temporal and spatial redundancy of adjacent frames in video data~\cite{xiang2020patchguard,xiang2021patchguard++}.

 To this end, focusing on the important live vision scenario that is widely adopted in autonomous driving and surveillance systems, this work proposes the real-time robust video object detection system, \textit{Themis}, to defend the adversarial patch attacks that practically introduce damaging consequences in video recognition tasks. We propose
 localized important superficial feature (LISF) based defensive methodology, which not only effectively identifies the adversarial regions but is also able to leverage temporal and spatial redundancy of video data for efficient robust object detection. 
 
 %To the best of our knowledge, this is the first systematic work targeting the practical attack models in live vision scenarios. 
 
 %we propose the robust video object detection framework, \textit{Themis}, with both algorithm and architecture innovation to effectively and efficiently detect the adversarial features and occlude them during video object detector inference to recovery system from adversarial effects.
 
  \emph{Themis Algorithm}\footnote{Themis, (Greek: “Order”) in Greek religion, personification of justice,  a \textbf{blindfolded} goddess holding a pair of scales.}: 
 We draw the key observation that adversarial inputs induce VOD to be overshadowed by the localized but inductive superficial features,  
 and the effect of adversarial patches can be eliminated facilely without aggravating the prediction accuracy of benign images when moving out LISFs. Hence, we propose LISF-based detection and recovery method based on prediction stabability testing by moving out LISFs. Results show that Themis algorithm effectively locates the adversarial regions and eliminate adversarial effects.

 \emph{Themis Architecture:}  Although \textit{Themis} algorithm is effective at detecting the adversarial regions in input data, it also introduces challenges due to the additional  multiple rounds of inference during the prediction stability testing by occluding LISFs. To reduce performance overhead to support real-time robust video object detection, we propose the \textit{Themis} architecture to eliminate both the inter-frame and intra-frame redundant computations. 1) Inter-frame: \textit{Themis} leverages the spatial and temporal redundancy of video data to eliminate  unnecessary computations in non-key frames. Specifically, for key frames of video data, the complete  defensive algorithm is performed to locate adversarial region or features. For non-key frames, by leveraging the temporal and spatial locality in video data, Themis approximately predicts the adversarial region location or features in non-key frames by estimating the motion movement of adversarial regions and features (i.e. optical flow in this work).
 2) Intra-frame: In observing the redundant computations of benign features during LISF-based detection and recovery, we propose the efficient hardware design for benign feature computation reuse and eliminate redundant unnecessary computation and memory traffic. Themis architecture can be
 easily integrated with existing DNN accelerators to friendly support state-of-the-art performance-oriented or accuracy-oriented video recognition methodologies
 %To our best knowledge, this is the first software/hardware system to efficiently defend against adversarial patch attacks during on-line inference.
In summary, this work has the following contributions:
%instead of masking every region of the original images to obtain the prediction decision map for adversarial patch detection~\cite{mrd}. \textit{Themis} avoid the  one's view of the important overshadowed by the trivial
 % with both image-based and feature-based warp strategies,
  %non-key frames.  
 %\textit{Themis} algorithm is capable of identifying the adversarial region of inputs. 
 %Therefore, it is compatible with video recognition framework that leverages both temporal and spatial redundancy. 
% Themis framework can be friendly integrated with state-of-the-art performance-oriented or accuracy-oriented video recognition methodologies {with light efforts}.
 %to support \textit{Themis} technique, so that \textit{Themis} have good applicability and usability.}

%contribute to address the problem of adversarial patches in 

\begin{itemize}
\setlength{\itemsep}{0pt}
\setlength{\parsep}{0pt}
\setlength{\parskip}{0pt}
  %\item We characterize feature space of benign and adversarial input data based on the superficial features, which proved  to have both good distinguishability and close spatial correlation to inputs.
  \item We propose the LISF-based methodology to accurately identify the adversarial region locations in input data. The detection methodology can effectively work under adaptive attack when an adversary has white-box knowledge of defensive approaches.  
  \item We propose the defensive framework that significantly reduces defending overhead in non-key frames by leveraging temporal and spatial locality in video data, which can be friendly integrated with the state-of-the-art video object detection frameworks. 
  
  \item We propose lightweight hardware customization to efficiently support the defensive framework and fully exploit the computation reuse of benign features, which can be easily adapted to existing deep learning accelerators. %and eliminating the redundant computations.
  
  \item We make an extensive experimental evaluation and the results show that \textit{Themis} system can effectively and efficiently defend adversarial attacks in real-time. Compared to the system without defensive mechanisms, \textit{Themis} improves the average mAP of real-time object detection from 0.03 to 0.66, with 1.08\% of hardware overhead. 
\end{itemize}

%\Note{Contribution summary}
%\subNote{algorithm: intra-frame, inter-frame}
%\subNote{arch design: intra-frame, inter-frame}
%\subNote{experiments results}

% \vspace{-5pt}
\section{Background}
% \vspace{-5pt}

\subsection{Video Object Detection Preliminary}
%\vspace{-5pt}

Video object detection,  recognizing instances of visual objects (e.g, humans, cars, animals) and their locations in digital videos, is a fundamental important computer vision task. It forms the basis of many other computer vision tasks, such as instance segmentation~\cite{ pinheiro2016learning}, object tracking~\cite{choi2012unified}, and image captioning~\cite{lu2018neural}, etc. 
%\noindent\textbf{Image Object Detection:}  pinheiro2015learning,

\noindent\textbf{Object Detection in a Single Image:}
Image object detection is the foundation of detecting objects in videos. Image object detector solves the following two subtasks: 1) Predicting how many objects are in the images. 2) Classifying these objects and estimating their locations with bounding boxes. The most important evaluation metric for object detector prediction accuracy is mAP (mean Average Precision) which considers both precision and recall rate. 
Recently, video object detection capability has been largely boosted by deep learning techniques with milestones of CNN-based image detectors, such as RCNN, YOLO~\cite{ redmon2017yolo9000}, SSD~\cite{liu2016ssd}, RetinaNet~\cite{lin2017focal}, etc. 
%These image object detectors  redmon2016you, , redmon2018yolov3

\noindent\textbf{Object Detection in Video Data:} Video data consists of many time-sequential images. Hence, video object detection can be achieved by performing image object detection in every image (referred to as accuracy-oriented (AO) schema). For better computing efficiency, prior video object detection methodologies are proposed to leverage temporal redundancy across frames in time-sensitive applications~\cite{dff,vrDANN,eva2} (referred to as performance-oriented (PO) schema).
{The key design concept of PO schema is to undergo the precise computation of  key frames, while approximately computing non-key frames based on key frame {features} and the motion, trajectory, or optic flow information.}

\begin{figure}[t!]
\centerline{\includegraphics[width=0.5\textwidth]{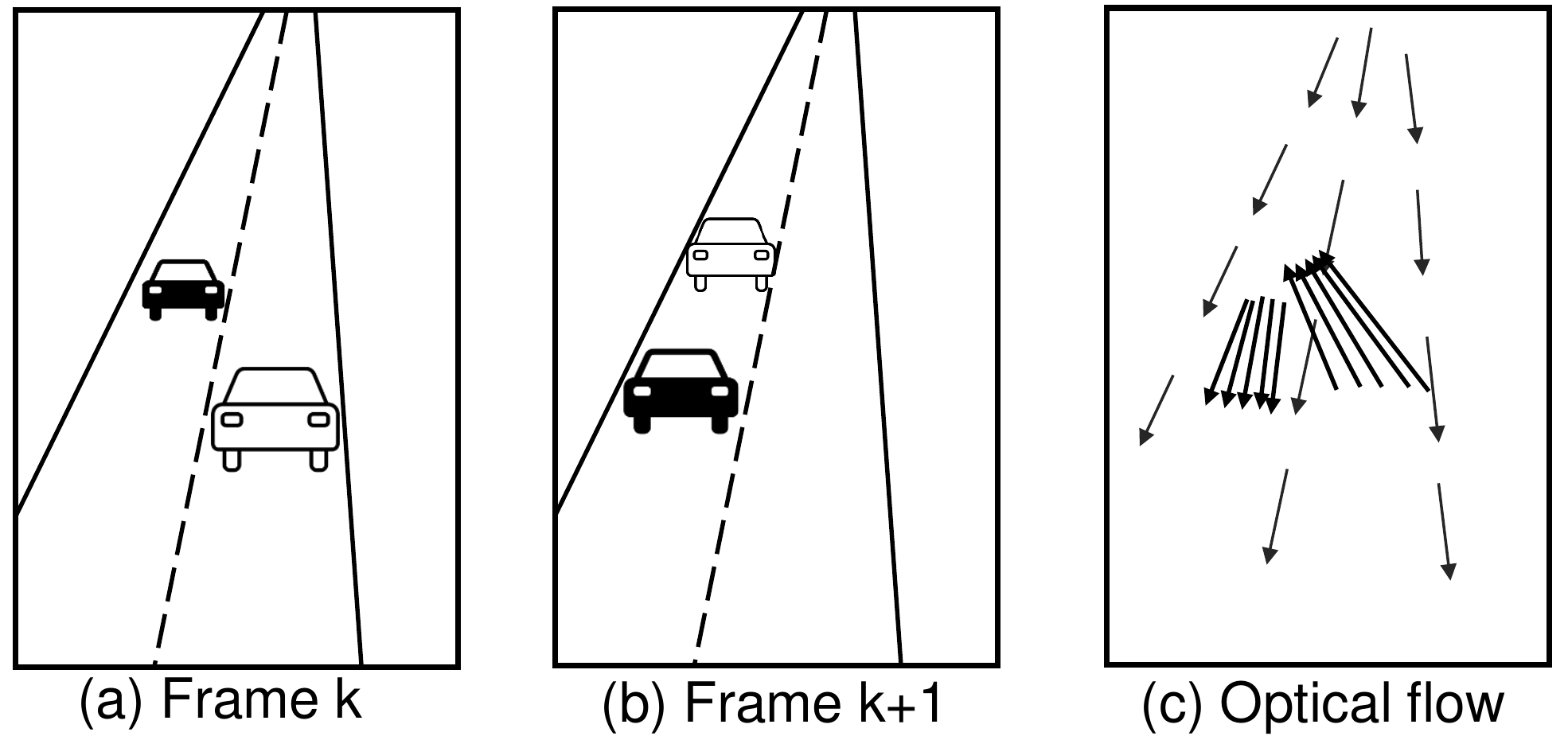}}
\vspace{-10pt}
\caption{Optical flow (2D vector field) describes the motion of objects in two frames.}
 \vspace{-10pt}
\label{fig:OpticalFlow}
\end{figure}

\noindent\textbf{Optical Flow:}
%Optical flow is one of the most fundamental problems in the field of computer vision, which can be widely applied in many applications, such as autonomous driving action recognition and robotics~\cite{zhai2021optical}.
Optical flow is widely used to utilize the temporal redundancy in video data and to approximate the non-key frames.%the approximation of non-key frames. 
Optical flow describes the apparent motion of image objects in consecutive frames caused by the movement of objects or the camera. Specifically, as shown in Fig.~\ref{fig:OpticalFlow}, optical flow(c) is a 2D vector field where each vector is a displacement (dx, dy), showing the movement of pixels from first frame (a) to second (b). 
Once we obtain the optical flow, the pixels or feature map of non-key frames can be estimated by warping the pixels or feature map of their predecessor key frame with optical flow: 
 %Further on, we can warp the feature map between those frames. The warp process can be defined as
$$
V(x+dx, y+dy, k+1) \triangleq V(x,y,k)
$$
where \textit{V} can be the pixel values or feature activation and $\triangleq$ can be implemented with different interpolation methods. The optical flow should be resized to warp the feature maps with different size.
Optical flow can be calculated based on pixel matching ~\cite{kroeger2016fast} or CNN ~\cite{ ranjan2017optical}, with the assumption that objects maintain the same intensity (brightness) between consecutive frames.
% lucas1981iterative, dosovitskiy2015flownet,

\begin{figure}[t!]
\centerline{\includegraphics[width=0.50\textwidth]{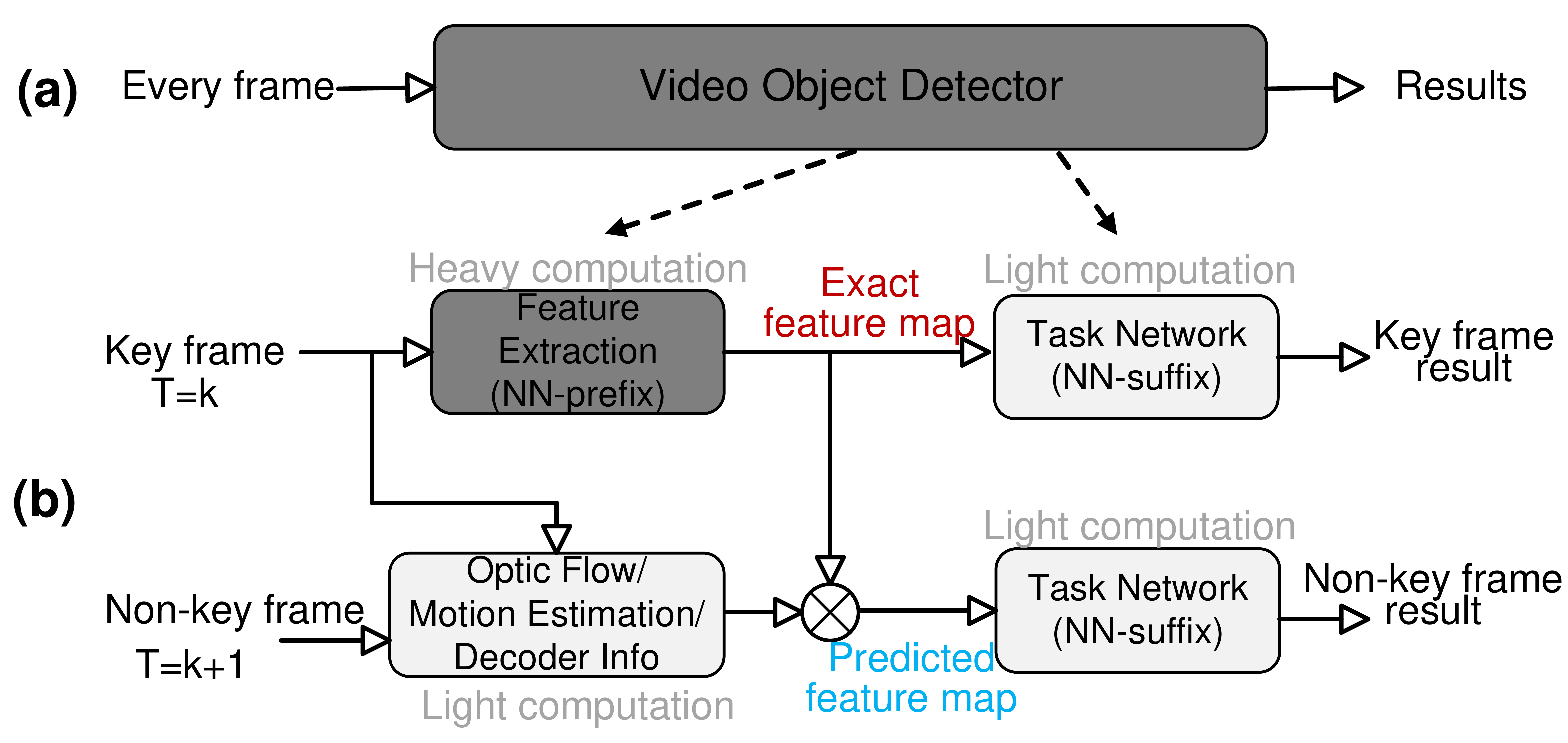}}
\vspace{-10pt}
\caption{Video recognition framework: (a) Accuracy-oriented (AO) framework precisely computes every frame; (b) Performance-oriented (PO) framework precisely computes key frames, while approximately computes non-key frame based on optical flow. }
\vspace{-5pt}
\label{fig:VRframework}
\end{figure}

\noindent\textbf{AO and PO Video Object Detection Details:}
{
Fig.~\ref{fig:VRframework} illustrates  accuracy-oriented (AO) and performance-oriented (PO) video object detection frameworks. For AO framework, the {image pixels} of every frame are input to the full video object detector for precise computation. }
{For PO framework, %, only key frames are input to the full video object detector for prediction. Non-key frames, on the other hand, reuse the key-frame features and skip the computation of feature extraction by taking the advantage of the spatial and temporal redundancy of video data across frames. Specifically, 
the DNN-based video object detector is divided into two parts: DNN-prefix for more generic feature extraction with much heavier computation and DNN-suffix for final prediction  with low computation overhead~\cite{eva2,asv}. PO framework first calculates the optic flow between non-key frames and the key frame.} % ,where optic flow is a form of visual streaming indicating objects moving continuously in one direction~\cite{dff}. 
% -----
{%Figure~\ref{fig:VRframework} illustrates  accuracy-oriented (AO) and performance-oriented (PO) video object detection frameworks. For AO framework, the {image pixels} of every frame are input to the full video object detector for precise computation. }
%{For PO framework, %, only key frames are input to the full video object detector for prediction. Non-key frames, on the other hand, reuse the key-frame features and skip the computation of feature extraction by taking the advantage of the spatial and temporal redundancy of video data across frames. Specifically, 
%the DNN-based video object detector is divided into two parts: DNN-prefix for more generic feature extraction with much heavier computation and DNN-suffix for final prediction  with low computation overhead~\cite{eva2,asv}. PO framework first calculates the optic flow between non-key frames and the key frame, where optic flow is a form of visual streaming indicating objects moving continuously in one direction~\cite{dff}. 
Then, warps the key-frame features with the resized and scaled optic flow information to approximately compute the non-key-frame features.  The predicted non-key-frame feature map is input to the DNN-suffix for the non-key frame result computation.} Due to the large gap between the computation of DNN-prefix and DNN-suffix, the computation overhead of non-key frames is significantly reduced. 
%warps: a numerical approximation operation based on the coordinate offset from optic flow
%skip the computation of heavy DNN-prefix, but take the advantage of the spatial and temporal redundancy of video data across frames, the PO framework can eliminate the heavy computation of feature extraction for non-key frames, but approximately compute its predicted feature map with the optic flow~\cite{dff}. Optical flow 
%Compared to the feature extraction, the latter operations are much lightweight and save significant computing overhead~\cite{asv}. The conceptual illustration of these two video object detection frameworks under accuracy-oriented and performance-oriented scenarios is shown in Figure~\ref{fig:VRframework}(a),(b). 
%Specifically, deep learning-based video object detection approaches can be divided into flow based with the assistance of optic flow net, tracking-based, motion-based with utilizing the video decoder information, LSTM (Long Short Term Memory)-based, attention-based, etc. 
These two VOD frameworks (image-based and feature-based) are adopted in different scenarios that are optimized for accuracy or performance. It is important to support effective and efficient defensive mechanisms in both these two cases. We propose \textit{Themis} framework to achieve this goal.
%that leverages the temporal and spatial locality of the video data to eliminate the defending overhead in both of these two VoD frameworks.  

%\textit{Themis} supports both two recognition frameworks with efficient defensive mechanisms that eliminate the defensive overhead in non-key frames.

%\textit{Themis} can be configured according to the accuracy or performance requirements. Some advanced studies leverage the time and spatial redundancy to address the low-quality issue of non-key frames. We also discuss the support for such video object detection frameworks in Section 7.1.

%adopt the flow-based methodology as the basic video recognition framework, which is one category of typical and classic approaches and performs good trade-off between accuracy and performance~\cite{??}. To note, our defensive approaches can also be extended to support other video detection techniques.

%\vspace{-5pt}

\subsection{Adversarial Attacks in Video Recognition }
%\vspace{-5pt}
\begin{comment}
\begin{figure}[t!]
\centerline{\includegraphics[width=0.5\textwidth]{fig/advT.pdf}}
\vspace{-10pt}
\caption{Real-world samples of adversarial attacks in video recognition~\cite{xu2020adversarial}: the person wearing the T-shirt printed with the adversarial patch is ignored by the detector.}
\vspace{-5pt}
\label{fig:advT}
\end{figure}
\end{comment}

\textbf{Attack Formalization.} Adversarial patch attack can largely damage video recognition tasks by evading video object detectors. It manipulates the victim model to output malicious results by adding the patch perturbation in the object of video data. Formally, the goal of adversarial patch attack is to generate the adversarial patch,  $\hat{P}$, to maximize the expectation of possibility for classifier $h$ to  output targeted malicious label $y_p$ with all adversarial inputs $x_p$ derived from data set $X$. 
%Adversarial patch attacks keep attract more attention because of its feasibility in real deployment scenarios. 
\begin{equation}
\hat{P} = arg  \max\limits_p E_{X}[logPr(h(x_{p})=y_p|{x_{p}})]
\label{Equation:patch}
\end{equation}

Patched image $x_{p}$ is generated by applying patch $p$ to $x$ in the input dataset $X$, which can be formalized as: %The attack goal is to generate the adversarial patch, $\hat{P}$, to maximize the expectation of possibility for classifier $h$ to  output targeted label $y_t$ with all adversarial inputs derived from data set $X$. 
%The poisoned input image data $x_{adv}$ with the adversarial patch on it can be denoted as :
\begin{equation}
    x_{p} = A(p, x), x\in X
\end{equation}
where $p$ is the adversarial patch, $x$ is the clean image, and A is the transformation function applying the adversarial patch on the clean image (environmental noises, resizing, rotations, and deformations). 

The adversarial patch determines the prediction results with a very small region of pixels (adversarial region) for a relatively broad range of input images. By moving out the adversarial region, the adversarial effects are eliminated and the prediction results are recovered. Hence, autonomously identifying the adversarial region in the video data is the key foundation of robust object detection. 
%\rebuttal{The adversarial patch attack can be generally formalized with Equation (1)-(2) with variants of different patch shapes, patch sizes, and patch transformations (environmental noises, rotations, and deformations)}. 

%by taking into account different kinds of transformations, such as environmental noises, rotations, and the object deformations. 

%\begin{equation}
%\hat{P} = arg  \max\limits_p %E_{X}[logPr(y_t|h({x_{adv}}))]
%\end{equation}

%\xing{explanation of the equation(3)}

\textbf{Patch Attacks vs. Example Attacks.}
Compared to \emph{adversarial example attacks}~\cite{fgsm} that have been largely studied in image classification tasks, patch attacks have the following advantages: 1) \emph{Better universality}, because the adversarial patch is independent of input images. Such a scene-independent feature enables physical-world attack without prior knowledge of the scene. Adversarial example attacks, on the other hand, generate perturbation noises highly dependent on the input images, which hinders their deployment in the physical world.  
2) \emph{Better robustness to environmental noises and geometric distortion}.  For example, human  being wearing the T-shirt printed with the adversarial patch can be ignored in the object detection systems in different environments and body gestures~\cite{xu2020adversarial}.  It is not only effective in a static figure input, but also in complex scenarios like walking, sitting, and running~\cite{xu2020adversarial}. 
More studies prove that adversarial patches are robust to not only environmental noises but also geometric distortion. 
%According to the statistics of the state-of-the-art attack study~\cite{xu2020adversarial}, the object detection rate can be reduced by  57\% in the physical world.

Adversarial example attacks, however, are scene-dependent and transfer poorly in different inputs. In real cases, the adversary neither can obtain the attack scene in advance nor compute the adversarial examples for every frame in real-time. Therefore, adversarial example attack is not a practical attack model in physical environments for video recognition tasks. In this following, we do not consider the adversarial example attacks.

\textbf{Limitation of Existing Defenses.} Although some prior researches propose the adversarial example detection methodology~\cite{deepFense,wang2020dnnguard}, the differences between adversarial patch and the adversarial example noises hinder these countermeasures' applicability in the patch attacks. 
%attacks are distinct from adversarial example attack which may hinder their applicability in the patch attacks.
Additionally, these countermeasures can only recognize whether the input is a benign image or not, but cannot recover the DNN system from the attack effect. In this study, we aim to rescue the video recognition tasks by not only detecting the malicious input but also eliminating the adversarial effect by moving out the malicious patches. More adversarial defense comparisons are introduced in Section~\ref{sec:relatedwork}. 

% \vspace{-5pt}

%\vspace{-5pt}
\section{ Adversarial Video Patch Characterization}
\label{sec:observation}
%\vspace{-5pt}

%In this section, we characterize the feature space of adversarial patches and draw the key observations for robust video object detection. 
Intuitively, patched images rely on the extremely localized important neurons in the adversarial regions to deceive and induce the object detector to output the incorrect results, while benign images perform stable object detection without relying on extremely localized important neurons. Hence, we perform the LISF characterization and have the following observations:

1) \emph{LISFs are good candidates for detecting adversarial regions in single frame.}

\begin{figure}[t!]
\centerline{\includegraphics[width=0.5\textwidth]{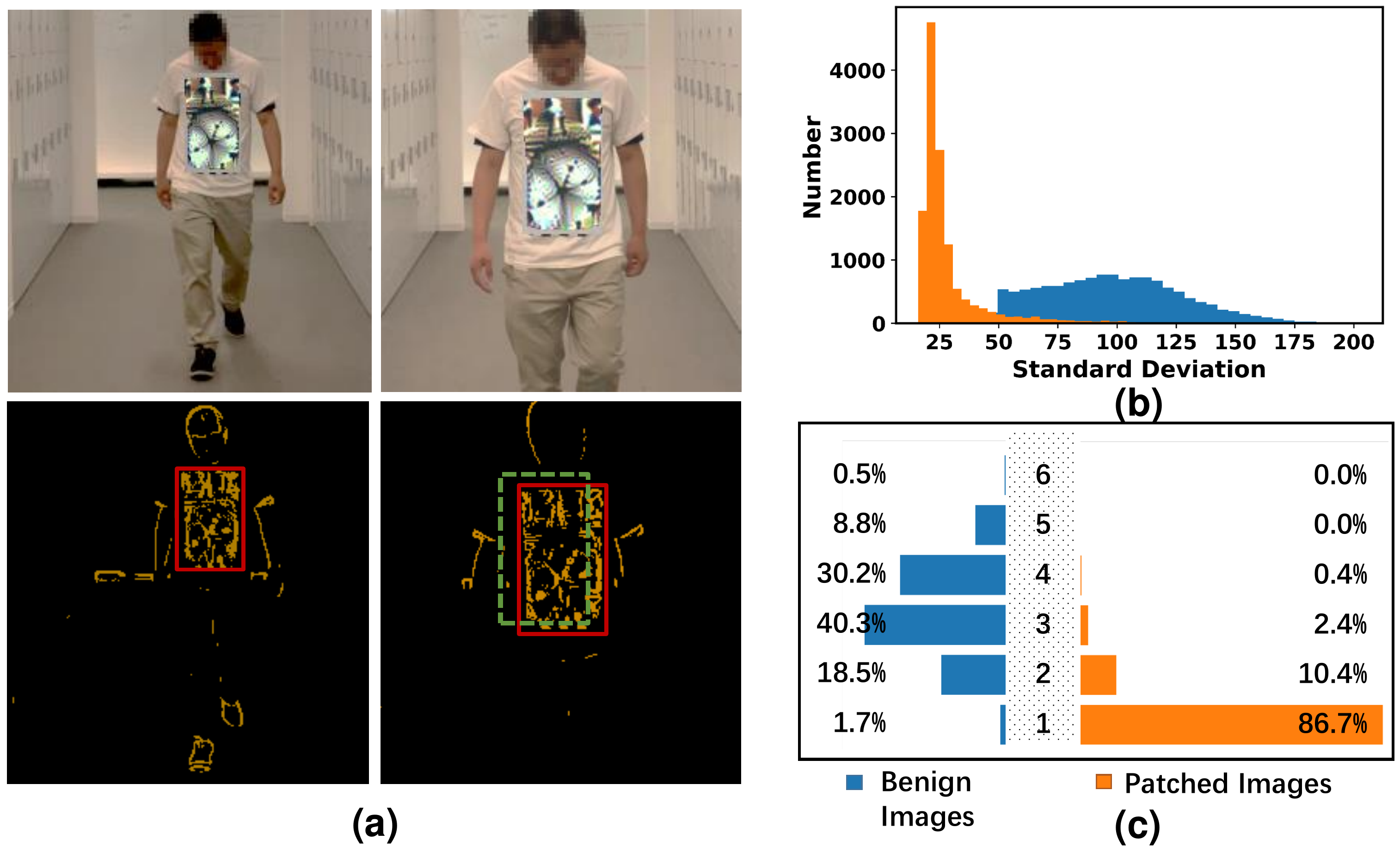}}
\vspace{-10pt}
\caption{Distribution of important neurons. (a) A showcase of the important neuron distribution and temporal association of LISFs along consecutive frames. 
%The patches are located by \textit{Themis} detection algorithm (boxes with the solid line) or estimated by optical flow (boxes with the dashed line). 
(b) The distance deviation of the important neurons in 12K images. (c) The average important neuron cluster number distribution in 12K images. }
\vspace{-5pt}
\label{fig:dist}
\end{figure}

LISF Distribution: The important neurons in feature maps of superficial layers exhibit a localized pattern in patched images while scattered in benign images. The superficial important features refer to the neurons that contribute significantly to the feature map value in the superficial layers (in this paper, we use the first layer).
Specifically, we take the superficial important neurons as the Top-K ones having the biggest value in the output feature map of the first layer. We visualize these important neurons of a patched image example in Fig.~\ref{fig:dist}(a). Intuitively, the patched image exhibits an extremely localized region of important neurons. Further, we make statistical counting about the distribution of Top-200 neurons of 12K images randomly selected from FLIC dataset~\cite{flic} based on two metrics: cluster distance and cluster number. We compare the standard deviation of the distance between the highlighted nodes to the central node in benign images and patched images. The patched images' distance deviation is much smaller than benign images, as shown in Fig.~\ref{fig:dist}(b). We also cluster the Top-k nodes with the classic MeanShift clustering algorithm. As shown in Fig.~\ref{fig:dist}(c), for patched images, about 86\% of cases have only one cluster and 97\% of cases have no more than two clusters. While for benign images, about 80\% of the cases have more than three clusters. Both of these two metrics show the localized distribution of superficial important neurons in patched images. 
 %We propose the  superficial input feature importance as the metric for discrimination analysis based on the intuition that in order to efficiently manage the output prediction results with a very small region of the input data, the adversarial patch must incur large activation from the first layer instead of the accumulation of the deep feature extraction.  More detailed comparison of superficial features and deep features (that widely used in related studies) is in Section~\ref{sec:featureimportance}. 

\begin{figure}[t!]
\centerline{\includegraphics[width=0.45\textwidth]{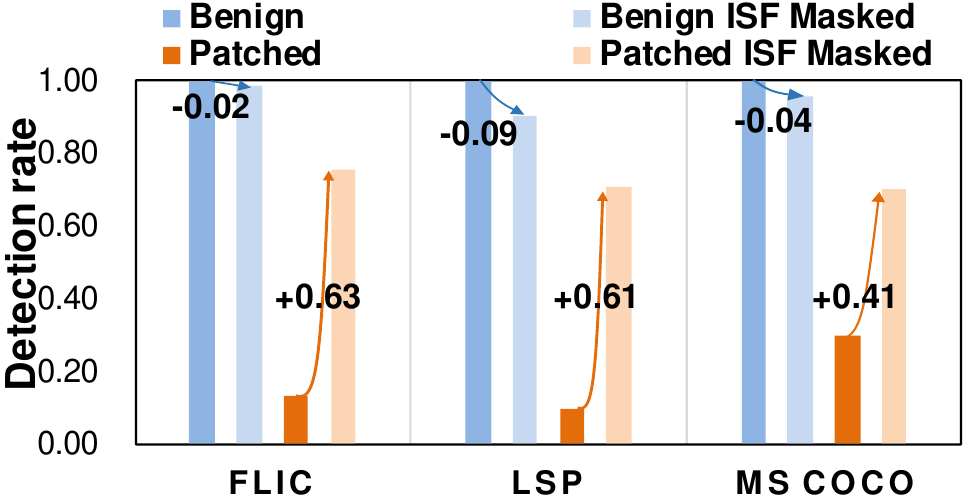}}
% \vspace{-5pt}
\caption{Prediction stability of occluding LISFs: benign inputs vs. patched inputs. }
\vspace{-5pt}
\label{fig:intra_stability}
\end{figure}

Prediction Stability by moving out LISFs: When occluding the localized superficial important features, patched images can be recovered mostly without affecting the prediction accuracy of benign images.  As shown in Fig.~\ref{fig:intra_stability}, we compare the detection rate of benign objects and patched objects before and after moving out the localized important features. The detection rate of benign objects decreases to 97.4\% slightly, while the detection rate of patched data increases from 17.9\% to 72.8\% significantly, which means that the prediction results of benign objects are much more stable than patched objects. 
For benign objects, the detection rate is not sensitive to the localized features. For the patched objects, patch effect would been effectively eliminated after moving the localized important superficial features.

2) \emph{LISFs exhibit temporal association in video data, which enables us to leverage temporal redundancy to eliminate the recovery overhead in non-key frames.}
%{When occluding the important neurons of key frames, the prediction results of benign video frames are much more stable than patched frames.} 
Superficial feature computing is very closed to the input, hence the important superficial features exhibit the similar temporal association as the image frames in video data. 
 The bottom two subfigures in Fig.~\ref{fig:dist} (a) show the LISFs marked within the solid red boxes in two consecutive frames.  The LISF in the following frames can be predicted by warping the optical flow  with the LISF in the previous key frame (predicted LISF marked in green box). 
 %Additionally, the adversarial effect is very sensitive to the occluding of LISF, 

In summary, LISF-based methodology not only effectively detects adversarial region and recovers the object detection in key frames, but is also able to leverage temporal redundancy in videos and eliminate defensive overhead in non-key frames. Based on these two observations, we then propose the LISF-based robust video objector detection system.

%adversarial region is inferred based on the LISF-based methodology. For the following consecutive frames, 

%\vspace{-5pt}
\section{Themis System Architecture}
%\vspace{-5pt}

%\vspace{-5pt}
\subsection{Overview}
%\vspace{-5pt}

\begin{figure}[t!]
\centerline{\includegraphics[width=0.50\textwidth]{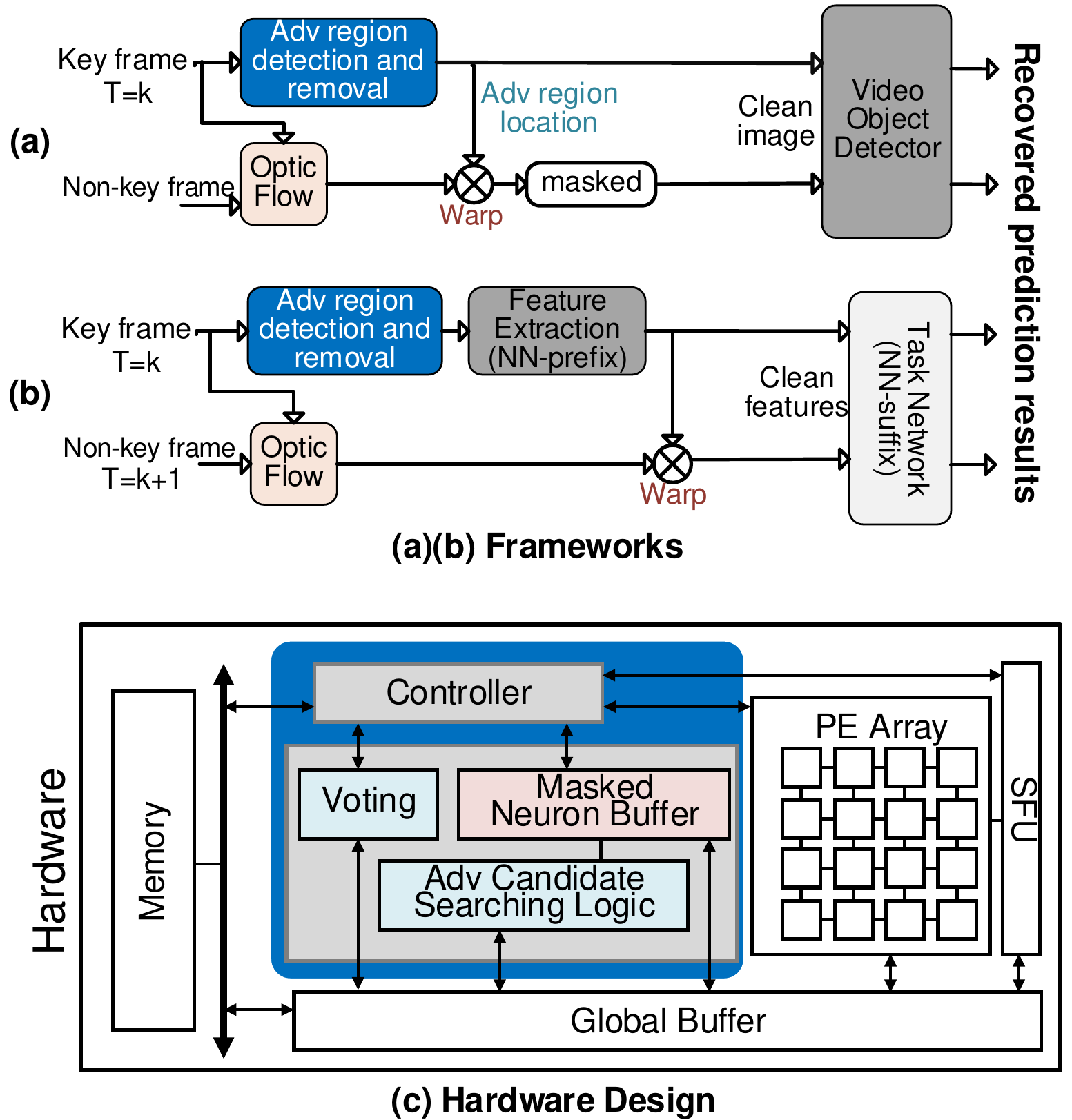}}
\vspace{-5pt}
\caption{Themis Architecture Overview. Themis reduce the defending overhead in non-key frames for both AO framework (a) and PO framework (b); Hardware optimization for benign feature computation reuse (c).}
\label{fig:sysdesign}
\vspace{-5pt}
\end{figure}

The overall \textit{Themis} system is shown in Fig.~\ref{fig:sysdesign} with algorithm, framework, and hardware designs. 
%The robust video object detection algorithm consists of three parts: video object detection, optical flow, and \textit{Themis} adversarial region detection (Section~\ref{sec:algorithm}). 
%To achieve real-time object detection and reduce the overhead introduced by adversarial patch defense, we propose the following system and architecture designs to eliminate redundant computations. 

\emph{Algorithms:} LISF-based methodology effectively targets the adversarial regions in input images and recovers the prediction results by moving out those regions.  

\emph{Framework:} Although the \textit{Themis} algorithm already reduces the adversarial region searching space with LISF-based methodology, multiple inferences are introduced in the occluding testing stage and incur large overhead. To achieve the goal of real-time robust object detection, \textit{Themis} proposes systematic design to efficiently remove redundant computations for better computing efficiency.

To further reduce the detection overhead, \textit{Themis} proposes the inter-frame and intra-frame optimizations to reduce redundant computations.  1) Inter-frame optimization: By leveraging the spatial and temporal locality between frames, \textit{Themis} only performs complete adversarial detection in key frames. For non-key frames, \textit{Themis} framework either predicts the adversarial region locations based on regions detected in key frames (image-based warping) or reuse the clean features in key frames after eliminating the adversarial effects (feature-based warping). These two warp strategies can be easily integrated with existing AO and PO video object detection frameworks.  % in adversarial features or the clean object features. 
2) Intra-frame optimization: The algorithm introduces a large volume of redundant calculations of benign features during occluding prediction stage for the key frames. Hence, \textit{Themis} scheduler reduces the computing overhead with computation reuse of benign features (Section~\ref{sec:scheduling}). Additionally, \textit{Themis} provides the interfaces of setting the following configurations for better feasibility: key frame proportion, the video object detector model segmentation strategy, and the configurable parameters in adversarial patch detection algorithms.

\emph{Hardware:} Although \textit{Themis} algorithm can be implemented in pure software, it is inefficient because of the following reasons: 1) Searching the heat-map of the input activation for the adversarial candidates is inefficient in the DNN accelerator due to the lack of the computing parallelism between the searching process and typical inference process; 2) During the voting stage,  multiple patch candidate regions will be occluded to calculate the inference results. In this process, a significant volume of benign features is computed repeatedly and introducing large unnecessary performance overhead. To address these issues, we propose the $Themis$ hardware architecture for efficient defense. The top-level block diagram of hardware architecture is shown in Fig.~\ref{fig:sysdesign} (c). Apart from the typical DNN accelerators with PE array, scalar function unit (SFU), global buffer, and control logic, \textit{Themis} is augmented with the LISF searching logic to search the candidate regions according to the heat map of the first superficial layer feature map, Masked Neuron Buffer to optimize the data and computation reuse of benign features, and the voting logic to decide the final prediction results.

%\vspace{-5pt}
\subsection{Framework}~\label{sec:scheduling}~\label{sec:warpStrategy}
%\vspace{-10pt}

\noindent\textbf{Inter-frame Optimization based on Tempo-Spatial Redundancy.}
The overview of $Themis$ framework is shown in Fig.~\ref{fig:sysdesign}(a) and Fig.~\ref{fig:sysdesign}(b) with support for AO and PO video object detection frameworks. Under both AO and PO scenarios, the defensive mechanisms for key frames are the same: $Themis$ algorithm  detects the adversarial region of frames and masks them to eliminate the adversarial effect. For non-key frames, $Themis$ proposes two warp strategies with  optical flow information which are friendly integrated in existing AO and PO video object detection frameworks, as shown in Fig.~\ref{fig:sysdesign}(a) and Fig.~\ref{fig:sysdesign}(b).  

1) \emph{Image-based Warping in AO Frameworks:} For AO framework, we approximately estimate the patch location in non-key frames by warping  the patch location in key-frame images with the optical flow. Specifically, every frame will be forwarded to the object detector for a complete inference. $Themis$ only performs the adversarial patch detection in key frames and obtains the adversarial region location. Then warps the detected adversarial region in key-frames with the optical flow information to estimate the adversarial region in non-key frames. The corresponding area of non-key frames is masked and the masked non-key frames are forwarded to the object detector for inference. The rationality is that the adversarial regions in the input data also exhibit temporal and spatial redundancy in live vision. Moreover, compared to the object to be segmented, the adversarial region is much less sensitive to the derivation brought by the inaccurate optical flow information (more validation in Section~\ref{sec:defensiveEffectiveness}). 

2) \emph{Feature-based Warping in PO Frameworks:} For PO framework, we directly warp the clean features in key-frames with the resized and scaled optical flow  to compute the non-key-frames features. Then, the predicted feature map is forwarded to the DNN-suffix for the final object detection results. Under both AO and PO cases, the defensive overhead of non-key frames is minimized.

\noindent\textbf{Intra-frame optimization for Benign Feature Computation Reuse.}
With the obtained coordinates of the patch candidates, \textit{Themis} then makes the prediction decisions by masking them from the original image individually. During this process, multiple inference rounds of inputs with different mask locations are introduced, which incurs a large execution overhead.  Hence, we propose the scheduling methodology with computation reuse to alleviate the execution overhead and eliminate redundant computations. As shown in Fig.~\ref{fig:MaskFlow}, the masked images and the original image share the most same pixels, with only the differences of masked regions. Thus, we compute the masked regions separately and splice the masked region back to the complete feature map for the final prediction results, to eliminate the most redundant computations. The detailed mask region computing is as follows: Given an image (224$\times$224 pixels) with the masked regions (50$\times$50 pixels). The pixels in the masked region are set as 0 to occlude their effect on the results. However, the features of the masked region through neural network layers are not simply set to 0, because the the existence of weight bias and the kernels that larger than 1. So we need to take the padding number into consideration to compute the features of the masked region.  For example, when computing the C1 layer in Fig.~\ref{fig:MaskFlow} for masked images, the complete region (masked region + padded region) for recomputation is 53$\times$53, with an additional purple part. The detailed computing process and dataflow optimization are in Section~\ref{sec:hwPatchSearch}.

\noindent\textbf{Reconfigurable Design Knobs.} \textit{Themis} framework provides the interfaces to configure the following design knobs that affect overall performance efficiency: the key frame proportion, the strategy of dividing the object detector to feature extraction DNN prefix and classification DNN suffix, the candidate numbers during adversarial patch detection. 

\emph{Key frame:} We set the key frame rate as 10\% with fixed length. \textit{Themis} framework can support the adaptive key frame strategies proposed by prior video object detection studies~\cite{eva2}. However, how to choose the key frame is out of the scope of this work. 

\emph{PO framework splitting:}
Designing DNN prefix and suffix in PO frameworks is the trade-off between performance and detection accuracy. When the prefix is close to the classification layers, the feature map is too small and may introduce larger accuracy degradation during optical flow estimation. We test extensive datasets and set the splitting spot when the feature map size is smaller than 56$\times$56. 

\emph{Adversarial region detection parameters:}
In the adversarial patch detection stage, the more adversarial candidates, the more rounds of the additional inferences and larger performance overhead are. The number of the patch candidates is determined by the heat spot selected threshold ($\beta$ and $\theta$).  We set $\beta = 0.75$ and $\theta = 0.85$. 

%\vspace{-5pt}
\subsection{Adversarial  Detection and Recovery}~\label{sec:algorithm}

 In observing that important superficial features exhibit different spatial distribution characteristics and exert distinct influence on the prediction results in benign and adversarial input data, we propose the LISF-based patch candidate searching methodology and then detect and eliminate the adversarial effect by occluding testing. 
 
\textbf{LISF-based Patch Candidate Searching.} We perform the LISF searching in the output feature map of the first layer. The size of the searching window is the upper limit of the patch sizes and the searching stride is 1. When the number of important neurons in one searching window is larger than the threshold ($\theta$), it is recognized as an important window and marked as the patch candidate.  When several important windows are overlapped, the central important window will be retained as patch candidate with all the other deleted. The detailed searching logic implementation is described in Section~\ref{sec:hwPatchSearch}.

\textbf{Occluding Testing and Recovery.} We recover the prediction results with the following two steps: 

1) \emph{Masked image execution}. After obtaining the candidate locations of adversarial patches, we generate masked images by occluding the patch candidate locations individually from the original images. These masked images are taken into the victim model to produce the prediction decisions. 
%After obtaining the candidate locations of adversarial patches, we make an inference by tearing them off the original images and then obtain the prediction decision individually. Then we identify the adversarial patches by checking whether there is a monopolist according to the inference results. 

2) \emph{Monopolist-occluded voting}. Themis performs the prediction decision analysis to detect patched images by examining whether there is a monopolist patch candidate that determines the prediction results. It is true that such a methodology will introduce the true-negatives. However, the results show that such cases are rare and \textit{Themis} can achieve good detection effectiveness. 
The detailed patch candidate searching and voting mechanisms are introduced as follows.  There are $k$ candidates: $P_1$, $P_2$, ..., $P_k$. The prediction result of the original image is $L_0$, and the corresponding prediction results of masked images: $L_1$, $L_2$, ..., $L_k$.  We first detect the patch by checking whether there is a $L_i$ distinct from others, while all the other labels are the same:
 If positive, the $P_i$ is the monopolist that dominates the prediction results, which is the adversarial patch. Only when tearing it off the image, the classifier predicts the robust and benign label $L_i$. 

If there is only one candidate, i.e. $k=1$, we compare $L_1$ and $L_0$ to determine whether there is an adversarial patch. If they are different, $P_1$ is the adversarial patch and $L_1$ would be recovered label. 

If there is no such particular label (either all the labels are the same, or several labels are different), no monopolist is detected. Then the image is recognized as a benign image. Themis then performs the majority voting to obtain the predicted label.

%\vspace{-5pt}
\subsection{Hardware Design}\label{sec:hwPatchSearch}
%\vspace{-5pt}

The Themis architecture can be integrated into the typical DNN hardware accelerator to support real-time detection with small overhead.  

\textbf{LISF searching logic.}
The LISF searching logic outputs the coordinates of clustered important neurons, which infers the possible candidate locations in the input image. LISF searching consists of the following three steps:

1) Obtain the binary important (output) feature map of the first layer. Computing Top-K neurons in the feature map is time-consuming and introduces large hardware overhead. Therefore, we make the estimation about the adaptive threshold according to the maximum value of the feature map. All neurons with activation values larger than the threshold ($\beta * feature_{max}$) are selected as important neurons. The important neurons map is stored in the buffer of LISF searching logic. 

\begin{figure}[t!]

\centerline{\includegraphics[width=0.5\textwidth]{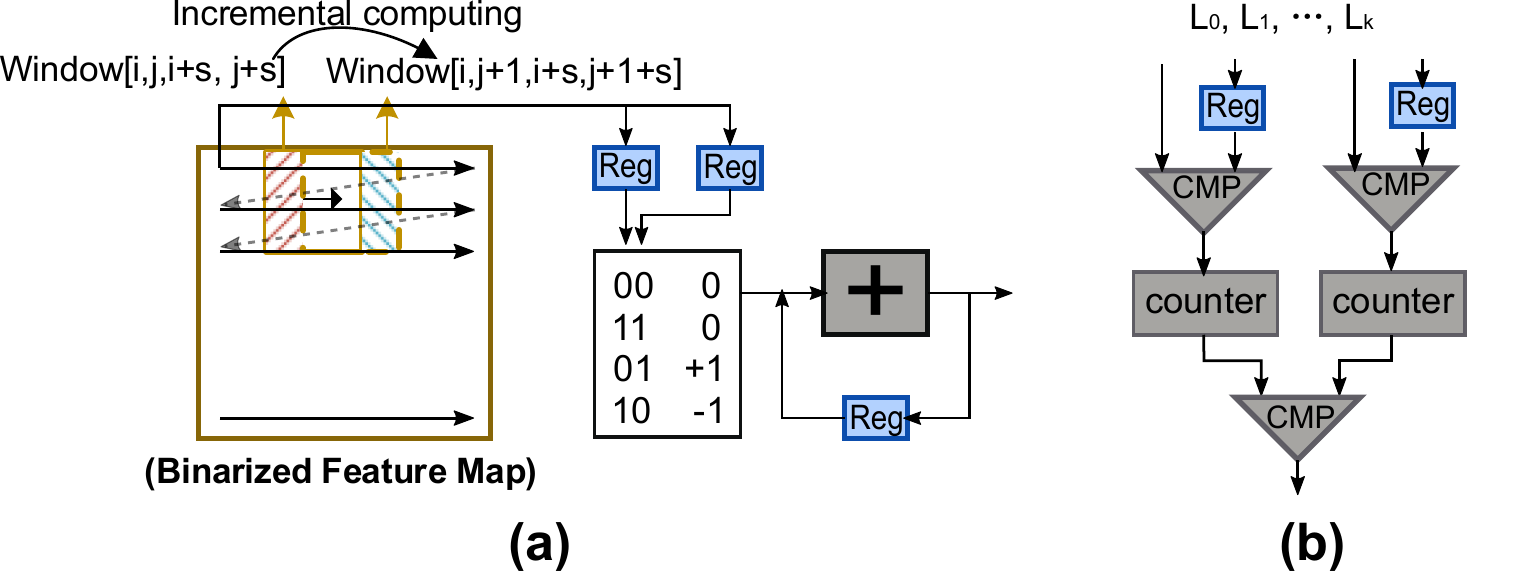}}
\vspace{-5pt}
\caption{Searching\&Voting logic.}
\label{fig:searchvoting}
\vspace{-5pt}
\end{figure}

2) Identify the important windows. We slide the fixed-size window to make statistic counting about the important neuron numbers in one window. When the important neuron numbers occupy more than the threshold ($\theta$) of the total neurons, we mark this window as an important neuron window, which will be highlighted as the patch candidates. To reduce the hardware overhead, we make the incremental accumulation of the important neurons in one window. As shown in Fig.~\ref{fig:searchvoting}(a), with the number of important neurons in Window[i,j,i+s,j+s], to compute the Window[i,j+1,i+s,j+1+s], we simply subtract the important neurons in column j+1 and add the important neurons in column j+1+s between row i and row i+s.

3) Delete the overlapped important windows. 
When two sliding important windows have more than 30\% of the area overlapped, we take them as one single patch candidate. 
% we treat them as a patch condidate, and keep the window with maximum important neurons as the final condidate and drop the other.

%To reduce the computation and memory access overhead, the sliding window statistical counting can be calculated differentially. As shown in Figure xxx.

Noted, the LISF searching is not on the critical path of DNN inference, which is parallelized with the processing of the original images.

\begin{figure}[t!]
\centerline{\includegraphics[width=0.50\textwidth]{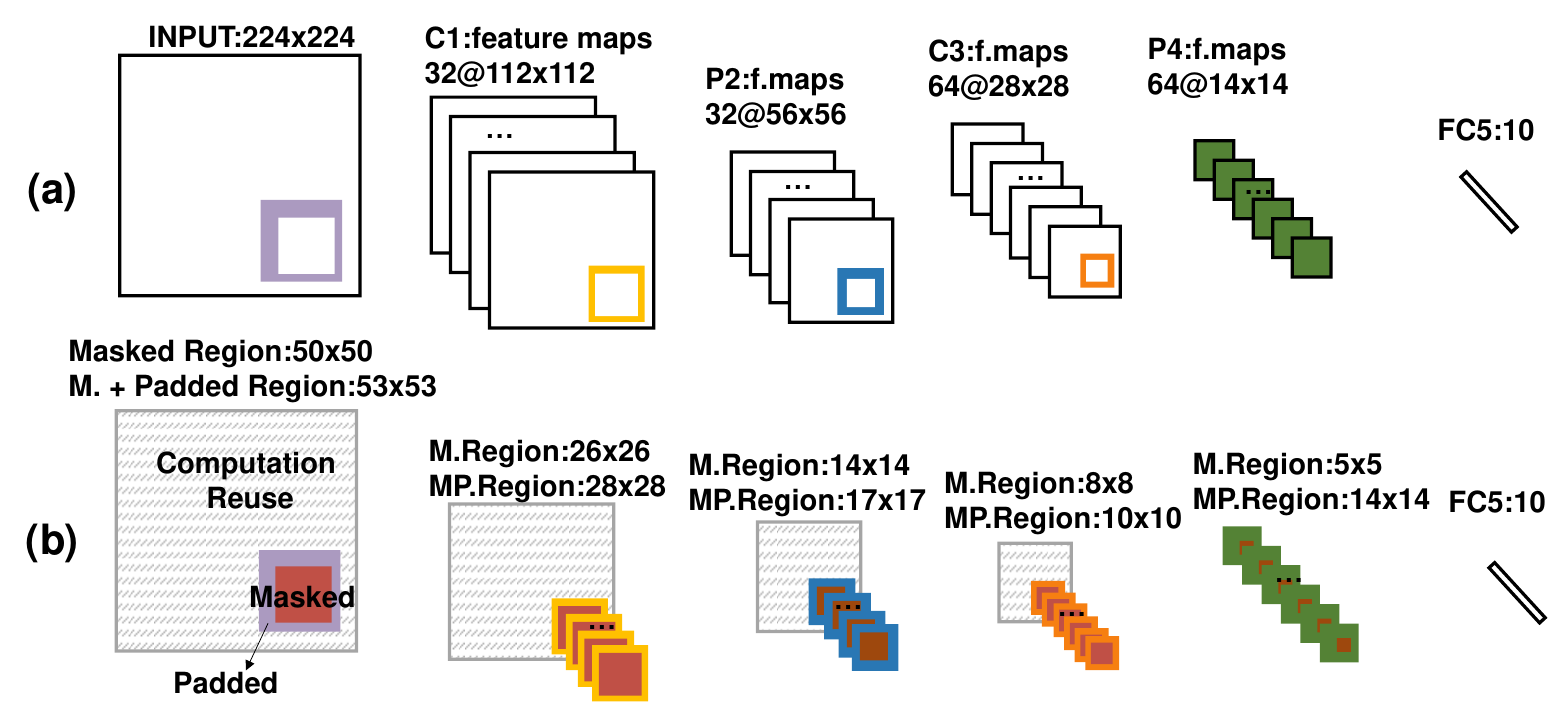}}
\vspace{-5pt}
\caption{Computing flow for original images and the masked images.}
\label{fig:MaskFlow}
\vspace{-5pt}
\end{figure}

\begin{figure}[t!]
\centerline{\includegraphics[width=0.50\textwidth]{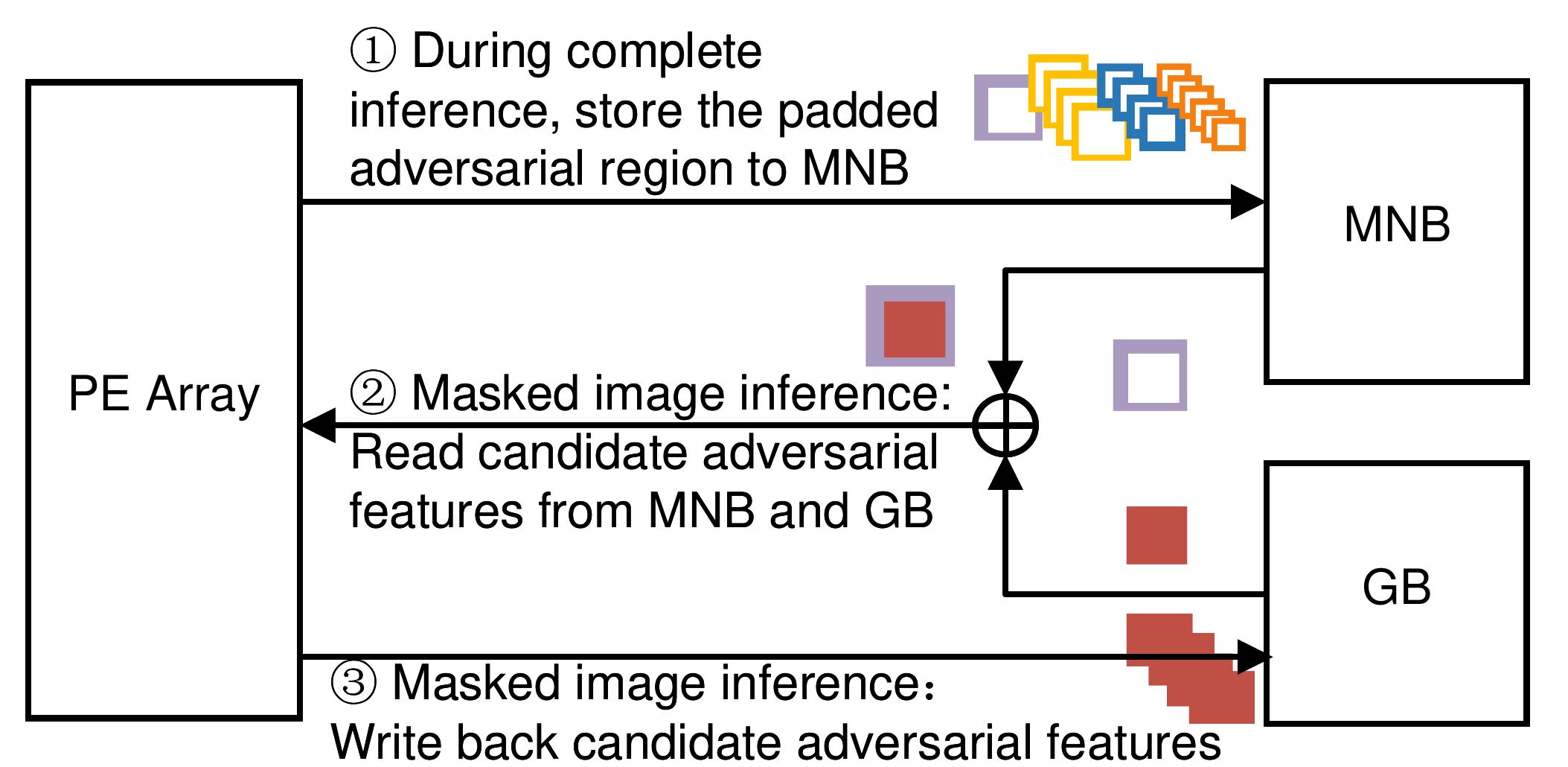}}
\vspace{-5pt}

\caption{Data reuse flow}
\label{fig:dataReuse}
\vspace{-5pt}
\end{figure}

\textbf{Masked Neuron Buffer (MNB).} To support efficient computation reuse of benign features and optimize the data accesses of the adversarial features, the masked neuron buffer is proposed to buffer the padded bounding area for feature computation of candidate adversarial regions. 

The data reuse flow of computing candidate adversarial regions is as follows (Fig.~\ref{fig:dataReuse}): With the coordinates of potential adversarial region candidates, the padded bounding areas through all the neural network layers in Fig.~\ref{fig:MaskFlow}(a) are determined. Through the typical inference, all the activation values of the padded bounding areas are stored in Masked Neuron Buffer. Then, during the inferences of masked images, all the masked regions are batched for computation. 
% For every layer, the PE array reads the candidate adversarial features from global buffers and the padded bounding areas from masked neuron buffers for the candidate adversarial feature computation. 
{For every layer, the PE array reads the candidate adversarial features from global buffers and the padded bounding areas from masked neuron buffers, as the red box and purple box shown in Fig.~\ref{fig:dataReuse}\textcircled{2}. The PE array combines these neurons by padding the red box with the purple box, takes it as input, and computes the candidate adversarial feature of next layer.}
After completing the computation of this layer, the adversarial features are stored back to global buffers.  Because the padded bounding area is very small compared to the full activation map, we configure the mask neuron buffer with a size of 8KB for every PE array.

\textbf{Voting Logic.}
The basic voting logic is as shown in Fig.~\ref{fig:searchvoting}(b). We use a compactor array to perform the pairwise comparison between the prediction labels of masked images ($L_0$, $L_1$, ...,  $L_k$). If there is an orphan label $L_i$, this image is identified as a patched image and the recovered label is $L_i$. Otherwise, it is a benign image and Themis performs the majority voting to obtain the recovered label. 

\textbf{Computing Flow.}
The overall computing flow is as follows: taking in a new input image, the DNN accelerator first performs the normal inference procedure on the image. Meanwhile, the searching logic obtains the candidate masked regions in the input based on the localized superficial important features. During the inference procedure of the original images, the feature data of corresponding padded bounding box are stored in the masked neuron buffer. After the completion of the original image inference, masked image inference rounds start to get their prediction results. After that, taking in those prediction results, voting logic detects the adversarial patches and output the recovered results.

For one input image, searching and voting operations are only performed once, while the inference rounds are determined by the searching candidates.Among these stages, multiple inferences introduce the most performance cost.  Voting is simple and performed within several cycles. Although the searching algorithm is time-consuming when offloaded to the CPU platform, its customized hardware largely boosts the performance and the overhead is less than 0.5\% of the multiple inferences. The detailed experimental evaluation is illustrated in Fig.~\ref{fig:breakdown}.

%Note{TO BE DETERMINED}

%We observe that the input share most of the same region, which provides the opportuntity of computation reuse. Then we propose the computation-reuse aware hardware accelerator to remove the redundant computations. 

%The detailed computing is as follows. 

%\vspace{-5pt}
\section{Experimental Methodology}
%\vspace{-5pt}

In the following sections, we will evaluate the defensive effectiveness and architecture efficiency of \textit{Themis}, which is complementary to enable the robust and real-time video object detection. 

%\vspace{-5pt}
\subsection{Validation on Algorithm Accuracy}
%\vspace{-5pt}

We test the attack success rate and the defensive effectiveness in both single-frame object detection tasks and the video object recognition tasks. For the previous scenario, we focus on exploring and validating the adversarial patch detection capability of \textit{Themis} algorithm on static images. In the latter scenario, we focus on exploring the defensive effectiveness on the non-key frames when \textit{Themis} framework leverages the temporal and spatial information in video data.       

\textbf{Attack Methodologies:}
For single-frame testing scenario, we adopt the digital-synthesized attack methodology that randomly attaches the digital adversarial patches onto the bounding box regions of the objects in MS COCO, FLIC, LSP datasets and random locations of images in ImageNet dataset with random rotated angles. The patch size is scaled with the area of bounding boxes ranging from 33x33 to 130x130 pixels. Since we focus on the defensive effectiveness of \textit{Themis}, we only perform attacks on the objects or images that can be correctly identified by the detector.  For video data testing scenario, we adopt physical attack videos released in 
the state-of-the-art attack methodology~\cite{xu2020adversarial} (Adv T-shirt), which significantly damages the functionality of YOLOv2 object detector. The patch size is variable and the adversarial patch has been significantly deformed during the human movement.

\textbf{Optical Flow Methodologies:}
\textit{Themis} framework is compatible with different optical flow methodologies. We use both the CV-based  and DNN-based optic flow methodologies: DIS~\cite{kroeger2016fast} and SpyNet~\cite{ranjan2017optical}, to validate the defensive effectiveness and architecture efficiency. The SpyNet is customized with scaled input image sizes and reduced pyramid levels. 

\textbf{Evaluation Metrics:} 
We adopt the commonly-used metrics, the detection rate and the standard mean average precision (mAP) score, to measure the accuracy of video object detection. In terms of performance efficiency, the frame per second (fps) is used to indicate how fast the framework process the video data. 

%\vspace{-5pt}
\subsection{Validation on Architecture Efficiency}
%\vspace{-5pt}

\noindent\textbf{ NN Accelerator Hardware Implementation.} Our methodology can be generalized adopted in diverse neural network accelerators. In this work, we evaluate our methodology based on the classic Eyeriss accelerator. Specifically, we augment the Eyeriss accelerator with the patch detector consisting of the LISF searching, masked neuron buffer, and voting logic. To prove the generality and its low overhead, we test both the server and edge accelerators with different configurations, as shown in Table \ref{table:config}. 

We implement the \textit{Themis} hardware design in Verilog RTL. To obtain the area and power, we synthesis and place\&route the RTL code with Synopsys toolchains under TSMC 28nm technology. We use CACTI 7 and DESTINY  to model the DRAM memory and on-chip SRAM buffers. Due to the unbearable long duration of silicon simulation, we also tailor an open-sourced simulator, nn-dataflow %\footnote{https://github.com/stanford-mast/nn\_dataflow}
to support \textit{Themis} for the total execution latency simulation. The simulator also reports the exact memory traces and module activities, which are then used to calculate dynamic energy consumption.

Besides, we deploy \textit{Themis} system on an off-the-shelf FPGA, Zynq UltraScale+ MPSoC ZCU104 board, to show the real performance. Specifically, we implement the proposed architecture components like Adv Candidate Search Logic and integrate them with a DPU (a soft IP for NN inference) using Vitis-ai framework.

\begin{table}[]
\caption{Hardware Platform Configurations}\label{table:config}
\vspace{-10pt}
\begin{center}
\scalebox{0.80}{
% \begin{threeparttable}
\begin{tabular}{|c|c|c|c|c|}
\hline
\multirow{2}{*}{\textbf{Mem}} & \textbf{Global Buffer/Array} & \textbf{RegisterFile/PE} & \textbf{Bandwidth} & \textbf{MNB\tnote{1}/Array}\\
\cline{2-5}             & 64KB                    & 256B                        & 26GB/s                     & 8KB              \\
\hline
\multirow{2}{*}{\textbf{Comp}} & \textbf{$\#$ of Arrays} & \textbf{$\#$ of PEs/Array} & \textbf{DataType}    &                   \\
\cline{2-4}              & 2$\times$2                     & 24$\times$24                            &  INT8                &                   \\
\hline
\end{tabular}
% \end{threeparttable}
}
\vspace{2mm}

\begin{tablenotes}
    \footnotesize
    \item[1] \textbf{MNB} denotes \textbf{M}asked \textbf{N}euron \textbf{B}uffer.
\end{tablenotes}
\end{center}
\vspace{-3mm}
\end{table}

% \vspace{-5pt}
\section{Experimental Results}
% \vspace{-2pt}

We first show the defensive effectiveness of $Themis$ in terms of single frame and video scenarios (Section~\ref{sec:defensiveEffectiveness}). Then we show the performance and energy efficiency of $Themis$ (Section~\ref{sec:PerfPower}), which introduces negligible overhead to real-time video object detection. Finally, we show that $Themis$ adds negligible area overhead to the baseline DNN accelerator(Section~\ref{sec:area}). 

%\vspace{-5pt}
\subsection{Defensive Effectiveness Evaluation}~\label{sec:defensiveEffectiveness}
% \vspace{-5pt}

We validate the defensive effectiveness of Themis under the following two scenarios: single-frame data and the video data with sequential frames.

\subsubsection{Single-frame Defensive Effectiveness}

\begin{figure}[t!]

\centerline{\includegraphics[width=0.5\textwidth]{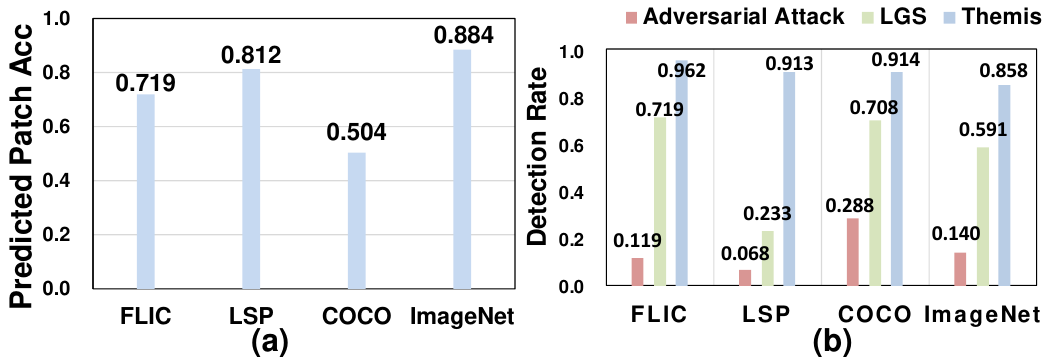}}
\vspace{-5pt}
\caption{Detection Effectiveness.}
\vspace{-5pt}
\label{fig:singlefmDefense}
\end{figure}

For single-frame testing, we use FLIC~\cite{flic}, LSP~\cite{lsp}, MS COCO~\cite{lin2014microsoft} and ImageNet~\cite{deng2009imagenet} datasets that are commonly used in object detection domain. We first evaluate the adversarial patch detection accuracy based on the metric of overlapped area proportion in Fig.~\ref{fig:singlefmDefense}(a). The predicted patch area and the actual patch area have an average of 72\%, 81\%, 50\% and 88\% overlapped region compared to the actual patch area size.  The results show that LISF-based searching methodology can accurately identify the location of adversarial patches. 
We then show the object detection rate before and after \textit{Themis} defense in  
Fig.~\ref{fig:singlefmDefense}(b) compared with Local Gradient Smoothing (LGS)~\cite{naseer2019local}  method. LGS locates the patch using local gradient of image pixels with the basic assumption that patch pixels are not smoothing. The detailed attack methodology is as follows: the adversary randomly attaches the adversarial patch in the person bounding box of the images in the datasets, so that the object detectors are evaded to ignore the persons.  With the adversarial patch attack, the object detection rate is 11.9\%, 6.8\%, 22.8\%, 14.0\% for FLIC, LSP, MS COCO and ImageNet. 
Compared to LGS that improves the detection rates to 71.9\%, 23.3\%, 70.8\%, 59.1\%.
, our \textit{Themis} defensive mechanisms works better and improves them to 96.2\%, 91.3\% , 91.4\% and 85.8\%.  The results show that \textit{Themis} can eliminate the adversarial patch effect effectively.

%%%Multiple patches%%%%\rebuttal{In the further step, we evaluate the detection rate under multiple patch scenarios. Specifically, we apply the patches on multiple objects (every object has a maximum of one patch) in images of FLIC dataset. The detection rate for 2 patches, 3 patches and 4 patches cases are 94.7\%, 91.3\%, 86.8\%. The high detection rates indicate that \textit{Themis} performs well even in challenging multiple patches cases.}

\begin{figure}[t!]
% \vspace{-5pt}
\centerline{\includegraphics[width=0.5\textwidth]{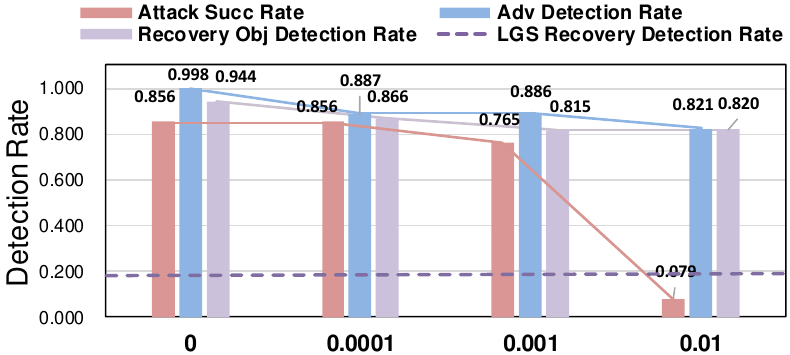}}
\vspace{-5pt}
\caption{Defensive Effectiveness under Adaptive Attack.The x-axis refers to the penalty coefficient $\alpha$.}
\label{fig:adaptiveAttack}
\vspace{-10pt}
\end{figure}

\textbf{Defensive Effectiveness 
Against Adaptive Attacks.} In the further step, we evaluate the defensive effectiveness under the strong adaptive attack~\cite{NEURIPS2020_11f38f8e} that the adversary gets known the full knowledge of the defensive strategies. To steer clear of the detection of $Themis$, the adversary trains the adversarial patch with Equation(3) that a penalty loss for the superficial activation value of patch is considered compared to Equation(1). In this way, the adversary aims to build the adversarial patch with good poisoning effects, but also trying to escape from the adversary candidate searching. $\alpha$ is the parameter to control the scale of the penalty loss in superficial activation value.  
\begin{equation}
% \vspace{-10pt}
% loss = -logit(y_p) + \alpha * \sum activation
% loss = -logit(y_p) + \alpha * \sum (w_1 * x_0)
% loss = -logPr(h(x_p)=y_p) + \alpha*\sum (w_1 * x_p)
loss = -logPr(h(x_p)=y_p) + \alpha*\sum (w_1 * patch)
%cost\left(X\right) = -log P\left( L^*|h\left(X\right)\right)
\label{Equation:loss_func}
\end{equation}

% \begin{equation}
% \hat{P} = arg  \max\limits_p E_{X}[logPr(h(x_{p})=y_p|{x_{p}})]
% \label{Equation:patch}
% \end{equation}

We perform the adaptive attacks on ImageNet dataset and the results are shown in Fig.~\ref{fig:adaptiveAttack}. Adversary detection rate refers to the rate that \textit{Themis} correctly identifies the adversarial region in the adversarial inputs or the benign input with no adversarial region. Recovery object detection rate refers to the rate that  \textit{Themis} correctly identifies the object in the image. The results show that, it is indeed that both the adversary detection rate and the recovery object detection rate decrease to 82.1\% and 82.0\% respectively, when $\alpha$ increases from 0 to 0.01. 
However, compared to the gentle slope of adversary detection rate and the recovery object detection rate, the attack success rate decreases much more drastically. When $\alpha$ is set to 0.01, the adversarial attack success rate is dropped to 7.9\%. As a comparison, after adding total variation loss, the detection rate of LGS drops signficantly from 59.1\% to 19.8\% while attack success rate maintains high.  These results indicate that the adversary cannot maintain the two goals of high attack success rate and good stealthiness simultaneously. \textit{Themis} can work effectively even under adaptive attack.

\subsubsection{Video Frame Defensive Effectiveness}

For the video frames, we adopt the 
adversarial video benchmarks in the state-of-the-art adversarial attack study~\cite{xu2020adversarial}, where the people wearing the adversarial T-shirt moving in indoor and outdoor scenarios and perform the practical attacks in the physical environment. The video object detector is based on YOLOv2. We evaluate the object detection rate and mAP under the following scenarios:
\begin{itemize}
\setlength{\itemsep}{0pt}
\setlength{\parsep}{0pt}
\setlength{\parskip}{0pt}
\item \emph{AO-ND}: AO framework with no defensive mechanisms.
\item \emph{AO-Full}: Defensive AO framework that examines every frames. 
\item \emph{AO-Dis}: Defensive AO framework that completely examines key frames, but CV-based methodology to predict the adversarial patch locations in non-key frame. 
\item \emph{AO-Spynet}: Defensive AO framework that completely examines key frames, but DNN-based optical flow information to predict the adversarial patch locations in non-key frame. 
\item \emph{PO-ND}: PO framework with no defensive mechanisms.
\item \emph{PO-Spynet}: Defensive PO framework with  DNN-based optical flow (spynet) . 
\end{itemize}

%The original video recognition accuracy with YOLOv2 for the original video datasets; 2) the video recognition accuracy with adversarial attacks; 3) the defensive video recognition accuracy with adversarial attacks. 

\begin{figure}[t!]
\vspace{-5pt}
\centerline{\includegraphics[width=0.50\textwidth]{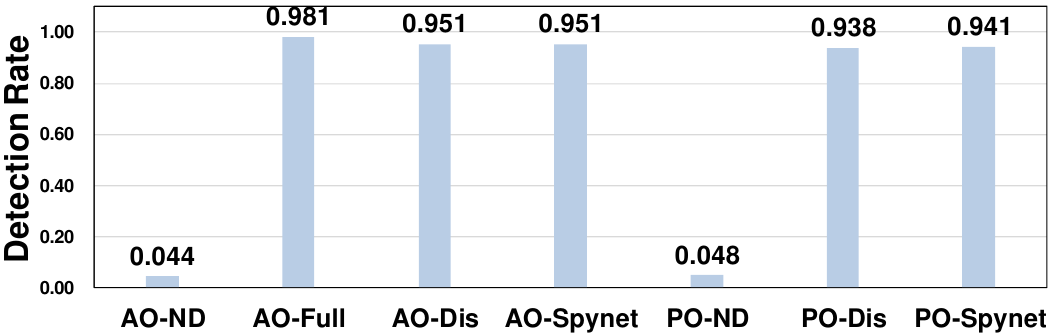}}
\vspace{-5pt}
\caption{Detection rate in video-frames.}
\label{fig:dr}
\end{figure}

\begin{figure}[t!]
% \vspace{-5pt}
\centerline{\includegraphics[width=0.50\textwidth]{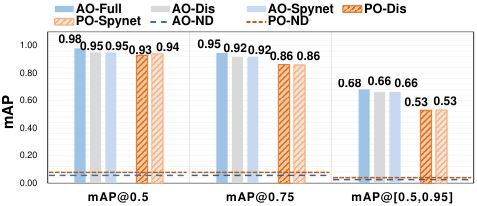}}
\caption{Defensive effectiveness in video-frames .}
\label{fig:mAP}
\vspace{-5pt}
\end{figure}

\textbf{Detection Rate:}
Fig.~\ref{fig:dr} shows the object detection rate under different scenarios. With adversarial attacks, both AO and PO object detectors have  significant low object detection rates of 4.4\% and 4.8\%, which indicates that adversarial attacks can essentially damage the integrity and functionality of object detectors even in the physical environments. With \textit{Themis} defensive algorithm, the detection rate is significantly improved above 93.8\% for different defensive strategies. Compared to AO-Full that examines every frame, other approximate defensive methods achieve relatively good object detection rates within a gap of less than 5\%. 

Detection rate is only a coarse-grained metric that intuitively indicates the detection recall rate of object detectors. In the further step, we evaluate the mAP that considers both the prediction accuracy, recall rate, and the predicted bounding box accuracy in the following. 

\textbf{mAP:} Fig.~\ref{fig:mAP} shows the mAP results under IoU = 0.5 (mAP@0.5), IoU = 0.75 (mAP@0.75), and average mAP value where IoU ranges from 0.5 to 0.95 with the step of 0.05. IoU (Intersections over Union) is the metric to determine whether it is an accurate prediction of the bounding box, which is calculated as the rate of dividing the area of overlap by area of union.  A larger IoU indicates a more strict criterion of mAP prediction accuracy. From the plot, specifically, we have the following observations:

1) Consistently, it is observed that adversarial attacks can effectively fool the object detector to ignore the human being with mAP as low as 0.03, 0.05 of AO-ND and PO-ND. With the guard of \textit{Themis}, the functionality of object detector is recovered and the average mAP is improved to the range of (0.53, 0.68) under different defensive mechanisms.  

2) When IoU is low (IoU=0.5), the mAP of defensive PO frameworks is equally good to the defensive approaches that examine every frame (AO-Full). When IoU is high, PO series may introduce the reduction of mAP due to the shift and deviation of the optical flow information.  Specifically, the gap between mAP@0.75 and the average mAP of PO-Spynet and AO-Full is 0.09 and 0.15.  

3) Adversarial patch location prediction is less sensitive to the deviation of optical flow. Although optical flow information also introduces the deviation between the predicted and actual adversarial regions, such deviation does not markedly hurt the defensive effectiveness of \textit{Themis} (less than 0.03), because of the prediction instability of adversarial regions, as analysis in Section~\ref{sec:observation}.

% \begin{figure}[t!]
% \vspace{-5pt}
% \centerline{\includegraphics[width=0.50\textwidth]{fig/Performance.pdf}}
% \vspace{-10pt}
% \caption{Performance comparison among different detection strategies.}
% \label{fig:performance}
% %\vspace{-10pt}
% \end{figure}

% \begin{figure}[t!]
% % \vspace{-5pt}
% \centerline{\includegraphics[width=0.50\textwidth]{fig/Energy.pdf}}
% % \vspace{-10pt}
% \caption{Energy comparison among different detection strategies.}
% \label{fig:energy}
% \vspace{-10pt}
% \end{figure}

%\vspace{-5pt}
\subsection{Architecture Efficiency Evaluation}~\label{sec:PerfPower}
\vspace{-10pt}

\subsubsection{Overall Performance and Energy Comparison}
\begin{figure}[t!]
% \vspace{-5pt}
% \centerline{\includegraphics[width=0.50\textwidth]{fig/Performance.pdf}}
\centerline{\includegraphics[width=0.5\textwidth]{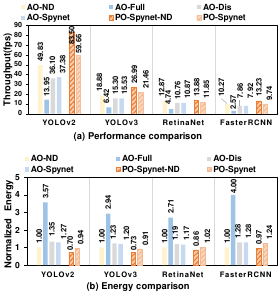}}
% \vspace{-5pt}
\caption{Performance and Energy comparison among different defensive strategies.}
\label{fig:performance_energy}
% \vspace{-5pt}
\end{figure}

We also evaluate the performance and energy of AO and PO frameworks with different defensive strategies. Fig.~\ref{fig:performance_energy}(a) shows the object detection throughput for four commonly-used video detectors: YOLOv2, YOLOv3, RetinaNet, and FasterRCNN. From the plot, we have the following conclusions:

1) PO frameworks significantly boost the performance when their model architectures can be divided into heavy NN-prefix and light NN-suffix. 

2) Performing adversarial detection in every frame incurs heavy overhead. Compared to AO-ND, AO-Full incurs 3.3x execution latency, which remarkably reduces the fps in all four object detectors. 

3) Eliminating the unnecessary recomputing for the adversarial region locations in non-key frames improves the performance and reduces the performance gap between defensive approaches with original approaches, while maintains the defensive effectiveness. Specifically, AO-SpyNet introduce 25.0\%, 17.7\%, 15.5\%, 22.9\% overhead compared to AO-ND. PO-SpyNet introduces 28.6\%, 20.4\%, 14.6\%, and 26.4\% overhead compared to PO-SpyNet-ND.

In summary, \textit{Themis} can effectively defend against adversarial attacks in video tasks, while still maintain the throughput of about 36 fps and 59 fps for real-time object detection in AO and PO frameworks.  

\subsubsection{Scheduling Optimization}

\begin{figure}[t!]
% \vspace{-5pt}
\centerline{\includegraphics[width=0.50\textwidth]{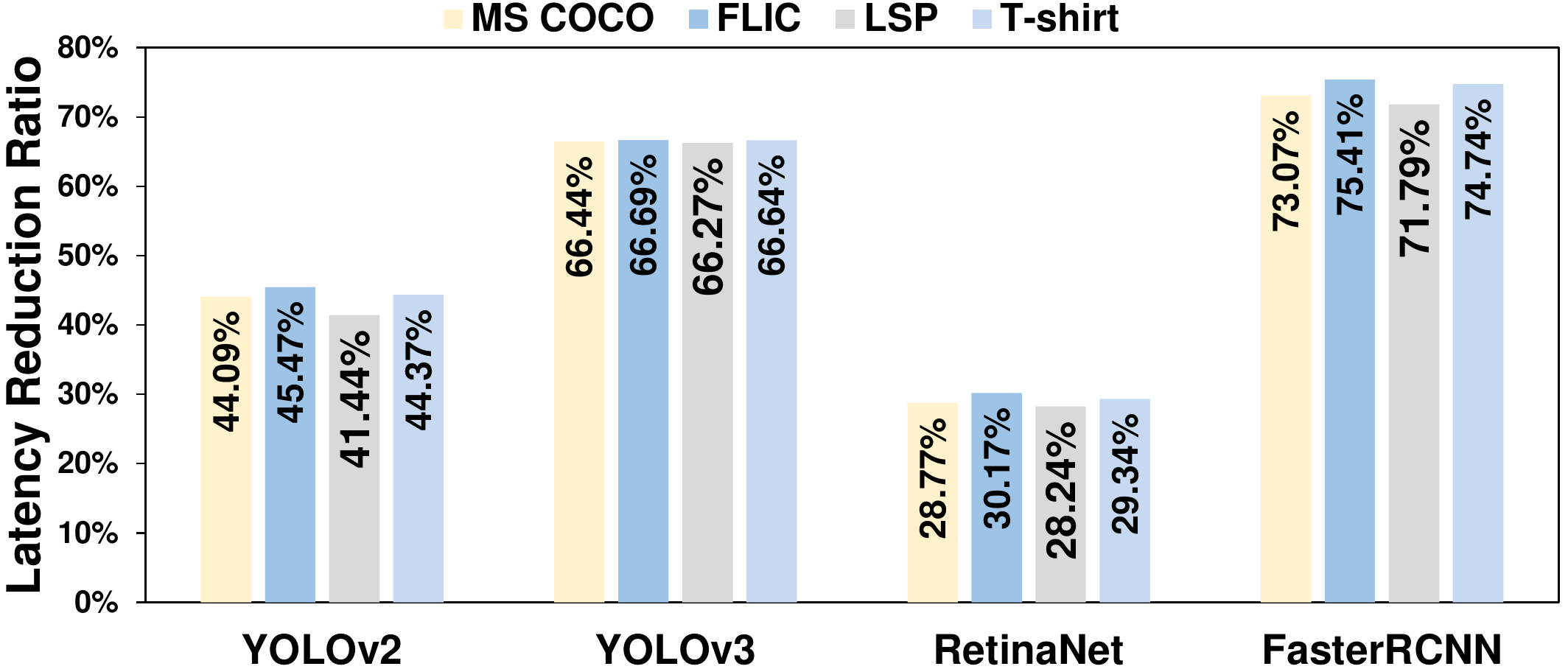}}
% \vspace{-5pt}
\caption{Latency reduction with benign feature computation reuse.}
\label{fig:computation_reuse}
% \vspace{-10pt}
\end{figure}

\begin{figure}[t!]
% \vspace{5pt}
\centerline{\includegraphics[width=0.5\textwidth]{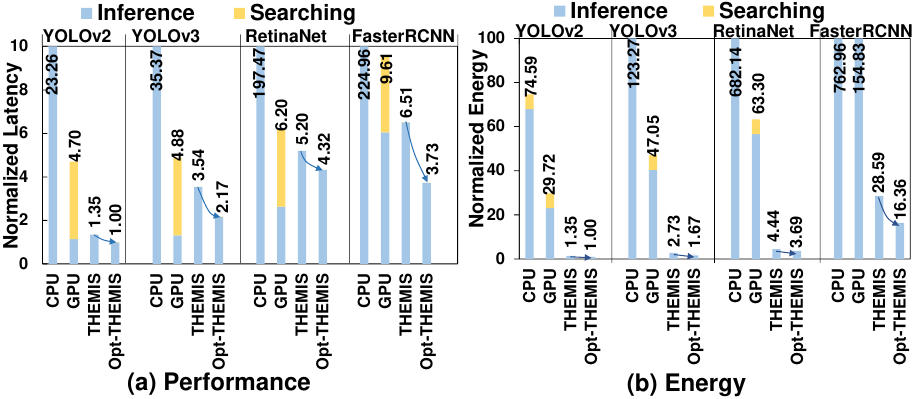}}
\vspace{-5pt}
\caption{Performance and Energy comparison with different architectures.}
\label{fig:breakdown}
\vspace{-5pt}
\end{figure}

We first evaluate the effectiveness of scheduling optimization for computing masked images in four typical datasets: MS COCO, FLIC, LSP and T-shirt~\cite{xu2020adversarial}. Fig.~\ref{fig:computation_reuse} shows the latency reduction ratio with benign feature computation reuse for different object detection models.
Patches on MS COCO, FLIC, LSP, and T-shirt account for 2.81\%, 1.47\%, 3.60\%, and 2.01\% of the whole image pixels on average, respectively.
We have the following observations:

1) \textit{Overall, Themis reduces the latency for masked images effectively with scheduling optimization.} The average latency reduction ratio for MS COCO, FLIC, LSP, and T-shirt are 51.93\% \textasciitilde 53.77\%, which indicates that the scheduling optimization in \textit{Themis} is applicable to various scenarios.

2) \textit{The scheduling optimization strategy in Themis is more efficient in object detectors with a shallower depth.} The average latency reduction ratio in FasterRCNN is 73.75\%,  which is significantly higher than 29.13\% in RetinaNet. The intrinsic reason is that the computation reuse ratios in the first several layers are higher than the latter. Specifically, for the first layer in YOLOv3, the computation reuse ratio reaches 98.38\%.

\textit{Themis} can efficiently reduce latency of the detection process for masked images with scheduling optimization. With an average latency reduction of 53.31\%, which significantly reduces the overhead incurred by the defense scheme.

\subsubsection{Comparison with different Architectures}

We compare the performance and energy of \textit{Themis} with CPU (Intel(R) Core(TM) i7-4770K at 3.50GHz) and GPU (NVIDIA TITAN V) platforms. Fig.~\ref{fig:breakdown} shows the normalized latency and energy for CPU, GPU and \textit{Themis} architectures running four video detectors: YOLOv2, YOLOv3, RetinaNet, and FasterRCNN with MS COCO dataset. 
$Themis$ hardware components can be integrated with any DNN architectures and we choose Eyeriss-like architecture as the basic DNN accelerator. Specifically, `Themis' refers to the cases of adopting candidate searching hardware components and `Themis-Opt' refers to the $Themis$ architecture with computation reuse optimization. For each case, we break down the system into inference procedure and searching procedure (CPU data bars are too large in this figure and are rescaled with the value marking on it). We draw the following conclusions: 1) $Themis$ is much more efficient compared to CPU and GPU platforms. Compared with CPU, $Themis$ achieves speedup from 16.3x to 60.3x. Compared with Titan V, $Themis$ reduces energy from 9.47x to 29.7x.
2) Candidate searching consumes non-negligible overhead in GPU platform. The customized searching logic achieves 745.7x speedup. 3) With the computation reuse optimization, Themis-Opt achieves speedup from 1.20x to 1.75x.

\subsubsection{Hardware Area Overhead}~\label{sec:area}
The baseline Eyeriss-like DNN accelerator has an area of 12.60 $mm^2$. On top of the basic DNN accelerator design, \textit{Themis} only introduces a total area overhead of 0.136 $mm^2$, which incurs only 1.08\% of hardware overhead. Specifically, the LFI searching logic occupies 0.0746\% of the hardware overhead, the masked neuron buffer occupies 1.005\%, and the rest is attributed to the voting logic. 
%\todo{update data}

% sum   eyeriss MNB     others
% 12.60 12.46   0.1266  3.13*0.00075*4

\subsubsection{Implementation on FPGA}
The experimental FPGA device (Xilinx  ZCU104 evaluation board) consists two ARM processor cores and 16nm FinFET+ programmable logic. The ARM processors are used to run Linux system and control the operation flow, while the programmable logic is reconfigured to accelerate the user applications.
\begin{comment}
\begin{figure}[t!]
\centerline{\includegraphics[width=0.5\textwidth]{fig/fpga2.pdf}}
\vspace{-5pt}
\caption{Experimental FPGA Device.}
\vspace{-5pt}
\label{fig:fpga}
\end{figure}
\end{comment}
We use the Vitis-ai framework to deploy \textit{Themis} with a customized DPU IP. Compared to the officially provided Yolov2 demo that runs at 24 FPS, \textit{Themis} achieves 22 FPS at AO defense mode and 35 FPS at PO defense mode. 
%\input{tex/results.tex}
%\input{tex/discussion.tex}

% \vspace{-5pt}
\section{Related Work}\label{sec:relatedwork}
%\vspace{-5pt}

\subsection{Real-time Video Object Detection}
%\vspace{-5pt}

Video object detection is the fundamental important computer vision task and grows rapidly with the increasing demands in autonomous driving and video surveillance~\cite{surveyVOD}.
Extended from image to video domain, DNN techniques largely boost video recognition capability. YOLO series~\cite{redmon2017yolo9000}, SSD~\cite{liu2016ssd}, RCNN series~\cite{ren2016faster}, are proposed in rapid succession.
Compared to single image object detection, video object detection has new attributes of existing both spatial and temporal correlations within consecutive frames. Previous studies leverage such attributes to improve the performance of video recognition tasks~\cite{dff, eva2,vrDANN}. %  add references here. fast:cvpr2016,
%Impression network~\cite{hetang2017impression} establishes the impression feature by iteratively absorbing sparsely extracted frame features to recognize objects from blurry frames. This impression mechanism makes it possible to perform long-range multi-frame feature fusion among sparse key frames with minimal overhead. 
%THP~\cite{zhu2018towards} introduced an adaptive key frame scheduling to improve the trade-off between speed and accuracy.
%THPM~\cite{zhu2018towards2} provides a lightweight network architecture for video object detection on mobiles.
%VR-DANN proposes the decoder-assisted DNN accelerator to leverage the motion vectors readily available in the decoding process and remove the tempo-spatial redundancy for real-time video recognition~\cite{vrDANN}. $EVA^2$ eliminates the temporal redundancy in live vision data by motion estimation that detects input changes and incrementally updates the old activation~\cite{eva2}. FAST leverages motion vectors in the compressed video to speed up super-resolution algorithms~\cite{fast:cvpr2016}. 
%FAST: A framework to accelerate super-resolution processing on compressed videos
In addition to the optimized video recognition frameworks that make use of temporal-spatial information for computing efficiency, some recent studies enhance feature maps with tempo-spatial information to improve detection accuracy that originally degraded by motion blur, rare poses, video defocus, etc~\cite{surveyVOD, FGFA}. {Besides, LSTM-based~\cite{liu2019looking},%, zhang2019modeling
attention-based~\cite{chen2020memory},%,deng2019relation,guo2019progressive,wu2019sequence,deng2019object
tracking-based~\cite{yang2019tracking}%,sharma2019online,kim2016cdt,feichtenhofer2017detect,
video object detectors are also proposed.
\textit{Themis} algorithm can support such video recognition frameworks well, because it is able to directly identify the adversarial region of input and supports feature aggregation or propagation operations during leveraging the tempo-spatial video information. }

%\vspace{-2pt}
\subsection{Defense Against Adversarial Attack}
%\vspace{-5pt}

%%%%%Rebuttal-Delete%%%%\textbf{Defense against adversarial patch attack.}
Adversarial patch attack is one of the most practical DNN attack models that can effectively damage object detectors even in physical environments~\cite{xu2020adversarial}. Envisioning its importance, prior studies contribute to the defense techniques against the adversarial patch attack by 1) building certified robust neural networks that resist the attack effect, such as IBP, de-randomized smoothing, etc~\cite{xiang2020patchguard,chiang2020certified,levine2020randomized}; %gowal2018effectiveness,cohen2019certified,zhang2020clipped  %like~\cite{chiang2020certified} proposes the first certified defense based on the interval bound propagation (IBP)~\cite{gowal2018effectiveness} that is commonly used in adversarial robustness certification problems. De-randomized smoothing (DS)~\cite{levine2020randomized} proposes the defense method extending randomized smoothing robustness schemes ~\cite{cohen2019certified} with structured ablation.
%which provide high confidence probabilistic robustness certificates. 
%Moreover, a provable robustness defense based on clipped BagNet~\cite{brendel2019approximating} with small reception field is proposed by~\cite{zhang2020clipped}.  
However, all of these certified studies achieve distinctly low certified accuracy for large scale datasets such as ImageNet. A certified accuracy of 20.5\%-36.3\% from these methods can hardly be applied in the real world systems.  
2) adversarial training to obtain robust model again patch attack~\cite{rao2020AdvTraing}, in which online re-training process introduces unacceptable cost.
3) performing patch detection based on empirical observations to eliminate the attack effect, such as digital watermarking (DW)~\cite{hayes2018visible}, local gradient smoothing (LGS)~\cite{naseer2019local}. % ,Inpaintint with Laplacian Prior(ILP)~\cite{anand2020ILP}
%For example, digital watermarking (DW)~\cite{hayes2018visible} utilizes the magnitude of the saliency maps to detect  unusually dense regions and mask them out of the input. Local gradient smoothing (LGS)~\cite{naseer2019local}  pre-processes the image gradient of the classification function with a normalization and a thresholding step and then suppresses the adversarial noise based on the gradient. 
However, these defenses proved to be invalidated when confronting with the strong adaptive attacker that has the white-box knowledge of the defense~\cite{chiang2020certified}. Note that since locality is the intrinsic property of patch attack, our LISF-based detection algorithm prevents the adversary from maintaining both strong attack effect and stealthiness, which promises the effectiveness of \textit{Themis} even under the adaptive attack.

The most related recovery methodology are MRD~\cite{mrd} and ObjectSeeker~\cite{xiang2022objectseeker}. MRD~\cite{mrd} requires a large amount of iterative inference passes to obtain the prediction map by masking every small region sliding across the entire original input images. It is extremely time-consuming and costs 1446\emph{s} for detection of one single image with sizes of 224$\times$224 on an Eyeriss-scale accelerator. % which is unacceptable for real-time detection.
Similarly, ObjectSeeker~\cite{xiang2022objectseeker} proposes patch-agnostic masking for certified objection detection that needs more than 100 inferences for different bands of the input image. Therefore, both MRD and ObjectSeeker are unacceptable for real-time detection due to their additional large cost.
%\textcolor{red}{Adversarial example detection methodologies~\cite{Ptolemy,dnnGuard,deepFense} are not applicable to the patch attack detection and recovery (more explanations in Section\ref{sec:patch})}\todo{guoqi: how about remove it?}. 

In summary, existing countermeasures are unable to perform online defense for video recognition tasks. This work proposes a high-efficient and effective detection and recovery system to defend the adversarial attacks that practically introduce damaging consequences in video scenarios. 

%\vspace{-5pt}
\subsection{Feature Importance Analysis}~\label{sec:featureimportance}
% \vspace{-10pt}

Neuron importance has been widely used for abnormal input detection~\cite{gan2020ptolemy,xu2020lance}%,xu2019dopa,li2021detecting
in previous studies. However, the metric (superficial feature importance) in our methodology is distinct from previous work. Previous studies focus more on the neurons that contribute significantly to the inference output (deep feature importance). We envision that deep feature importance is not a good candidate from the following two aspects:  1) both the benign images and the adversarial images have the deep feature importance. Its discrimination in the benign images and adversarial images is not straightforward, so that it requires more complex computation to identify the benign and adversarial images. 2) calculating such deep feature importance is time-consuming, which demands the gradient information and the complete backward propagation process. We propose the  superficial input feature importance as the metric for discrimination analysis based on the intuition that in order to efficiently manage the output prediction results with a very small region of the input data, the adversarial patch must incur large activation from the first place instead of the accumulation of the deep feature extraction.

%\vspace{-5pt}
\section{Conclusion}
%\vspace{-5pt}

\emph{Themis} efficiently and accurately recovers the DNN systems from the adversarial attacks with both algorithmic framework and the architectural support. At the algorithmic level, \emph{Themis} prevents the classifier from being overshadowed by the trivial but extremely biased parts by tearing the patch off the original images. At the architectural level, \emph{Themis} not only proposes efficient searching and voting logic, but also proposes the scheduling methodology to accelerate the masked image execution by eliminating the redundant computations and memory traffics. 
The results show that the proposed methodology can effectively recover the VOD system from the adversarial effect in real-time.
%%%%%%% -- PAPER CONTENT ENDS -- %%%%%%%%

% %%%%%%%%% -- BIB STYLE AND FILE -- %%%%%%%%
% \bibliographystyle{IEEEtranS}
% \bibliography{refs}
% %%%%%%%%%%%%%%%%%%%%%%%%%%%%%%%%%%%%

% \bibliographystyle{plain}
\bibliography{main}

%%%%%%%%%% -------- biography ----------- %%%%%%%%%%%%%%

\begin{IEEEbiography}[{\includegraphics[width=1in,height=1.25in,clip,keepaspectratio]{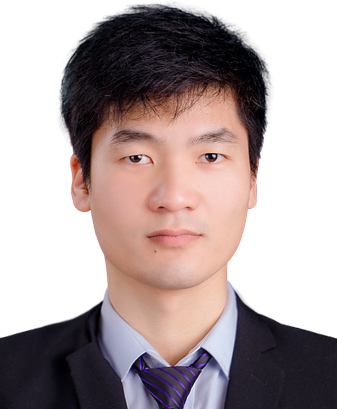}}]{Husheng Han}
% \begin{IEEEbiographynophoto}{Husheng Han}
received the B.S degree from Tsinghua University, Beijing, China in 2020. 
Currently he is working toward the PhD degree in the Institute of Computing Technology, Chinese Academy of Sciences, Beijing, China, and the University of Chinese Academy of Science, Beijing, China.
His current research interests include machine learning security and domain-specific hardware architectures.
\end{IEEEbiography}
\vspace{-30pt}

\begin{IEEEbiography}[{\includegraphics[width=1in,height=1.25in,clip,keepaspectratio]{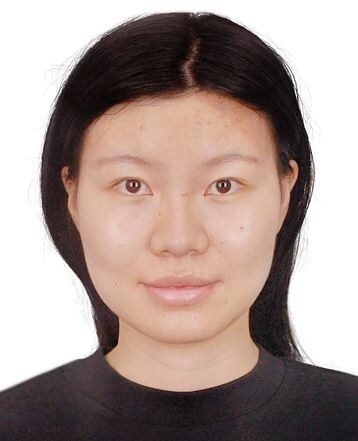}}]{Xing Hu}
% \begin{IEEEbiographynophoto}{Xing Hu}
received the B.S. degree from Huazhong University of Science and Technology,
Wuhan, China, and Ph.D. degree from University of Chinese Academy of Sciences, Beijing,
China, in 2009 and 2014, respectively. She is currently an associate professor of State Key
Laboratory of Processors, Institute of Computing Technology (ICT), Chinese Academy
of Sciences (CAS), Beijing, China. Her current research interests include domain-specific hardware architectures and deep learning system.
\end{IEEEbiography}
\vspace{-30pt}

\begin{IEEEbiography}[{\includegraphics[width=1in,height=1.25in,clip,keepaspectratio]{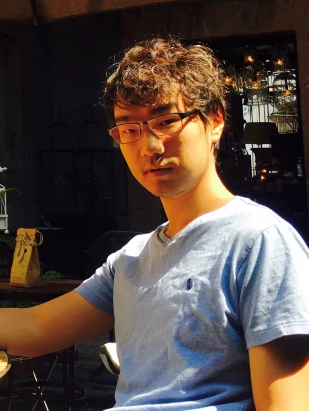}}]{Kaidi Xu}
% \begin{IEEEbiographynophoto}{Kaidi Xu}
received the B.S. and M.S. degrees from Sichuan University in 2015 and the Department of Computer Science at University of Florida in 2017 respectively and Ph.D. degree at Northeastern University in 2021. He is currently an Assistant Professor in Department of Computer Science at Drexel University, Philadelphia, USA.
His primary research interest is the robustness of machine learning, including physical adversarial attacks, rigorous robustness verification and certified defenses. 
\end{IEEEbiography}
\vspace{-30pt}

\begin{IEEEbiography}[{\includegraphics[width=1in,height=1.25in,clip,keepaspectratio]{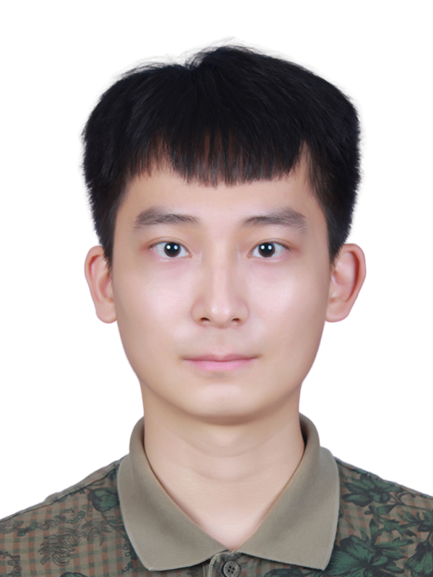}}]{Pucheng Dang}
% \begin{IEEEbiographynophoto}{Pucheng Dang}
received the B.S degree from Harbin Institute Of Technology, Harbin, China in 2019. 
Currently he is working toward the PhD degree in the Institute of Computing Technology, Chinese Academy of Sciences, Beijing, China, and the University of Chinese Academy of Science, Beijing, China.
His current research interests include machine learning security and Computer Vision.
\end{IEEEbiography}
\vspace{-30pt}

\begin{IEEEbiography}[{\includegraphics[width=1in,height=1.25in,clip,keepaspectratio]{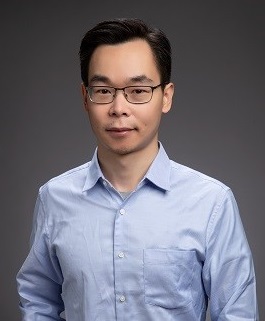}}]{Ying Wang}
% \begin{IEEEbiographynophoto}{Ying Wang}
% \todo{Add education.}
received the B.S and M.S. degree from School of Astronautics, Harbin Institute of Technology, Heilongjiang, China in 2007 and 2009 respectively and the PhD degree from the Institute of Computing Technology, Chinese Academy of Sciences, Beijing, China in 2014.
He is currently an Associate Professor in State Key Laboratory of Computer Architecture at Institute of Computing Technology, Chinese Academy of Sciences. 
His research interests primarily focus on the area of reliable computer architecture and VLSI design, with an emphasis on memory systems, energy-efficient accelerators, and approximate/error-tolerant computing. 
\end{IEEEbiography}
% \vspace{-20pt}

\begin{IEEEbiography}[{\includegraphics[width=1in,height=1.25in,clip,keepaspectratio]{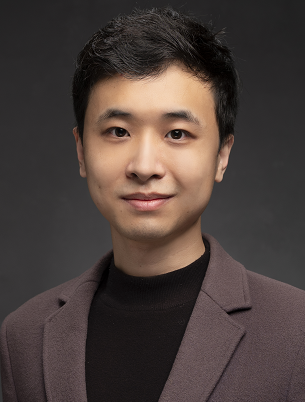}}]{Yongwei Zhao}
% \begin{IEEEbiographynophoto}{Yongwei Zhao}
% \todo{Add interests}
recieved the bachelor’s degree in computer science
and technology from the Huazhong University of Science and Technology, Wuhan, China, in 2015 and the PhD degree from the institute of Computing Technology, Chinese Academy of Sciences, Beijing, China, in 2020. He is currently an assistant professor at the Institute of Computing Technology (ICT), Chinese Academy of Sciences (CAS), Beijing, China.
His research interests primarily focus on the area of accelerator architecture.
\end{IEEEbiography}
\vspace{-20pt}

\begin{IEEEbiography}[{\includegraphics[width=1in,height=1.25in,clip,keepaspectratio]{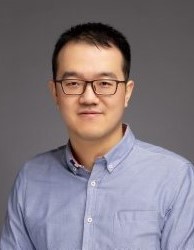}}]{Zidong Du}
% \begin{IEEEbiographynophoto}{Zidong Du}
(Member, IEEE) received his B.E. degree from the Department of Electronic Engineering, Tsinghua University, Beijing, China, in 2011, and the Ph.D. degree from the Institute of Computing Technology (ICT), Chinese Academy of Sciences (CAS), Beijing, in 2016. He is currently an associate professor at Intelligent Processor Research Center, Institute of Computing Technology (ICT), Chinese Academy of Sciences (CAS). His research interests mainly focus on novel architecture for artificial intelligence, including deep learning processors, inexact/approximate computing, neural network architecture, neuromorphic architecture. He has published over 20 top-tier computer architecture research papers, including ASPLOS, MICRO, ISCA, TC, TOCS, TCAD. For his innovative works on deep learning processors, he won the best paper award of ASPLOS’14, Distinguished Doctoral Dissertation Award of CAS (40/10000), Distinguished Doctoral Dissertation Award of China Computer Federation (10 per year).
\end{IEEEbiography}
\vspace{-20pt}

\begin{IEEEbiography}[{\includegraphics[width=1in,height=1.25in,clip,keepaspectratio]{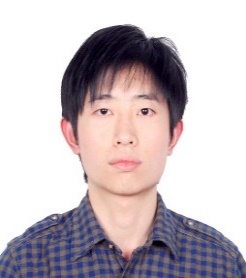}}]{Qi Guo}
% \begin{IEEEbiographynophoto}{Qi Guo}
(Member, IEEE) received the B.E. degree in computer science from Tongji University, Shanghai, China, in 2007, and the Ph.D. degree from the Institute of Computing Technology, Chinese Academy of Sciences, Beijing, China, in 2012.,From 2012 to 2014, he was a Staff Researcher at IBM Research, Beijing. From 2014 to 2015, he was a Postdoctoral Researcher with Carnegie Mellon University, Pittsburgh, PA, USA. He is currently a Professor with the Institute of Computing Technology, Chinese Academy of Sciences. His research interests include computer architecture, programming models, and machine learning.
\end{IEEEbiography}
\vspace{-20pt}

\begin{IEEEbiography}[{\includegraphics[width=1in,height=1.25in,clip,keepaspectratio]{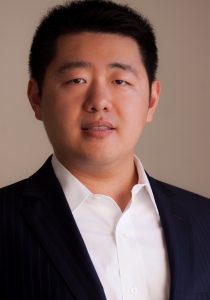}}]{Yanzhi Wang}
% \begin{IEEEbiographynophoto}{Yanzhi Wang}
received his B.S. Degree in Electronic Engineering from Tsinghua University, Beijing, China, in 2009 and the Ph.D. Degree in Computer Engineering from University of Southern California (USC) in 2014, under the supervision of Prof. Massoud Pedram.  
He is currently an Associate Professor and Faculty Fellow in the Department of Electrical and Computer Engineering, and Khoury College of Computer Science (Affiliated) at Northeastern University.
His research interests include real-time and energy-efficient deep learning and artificial intelligence systems, model compression of deep neural networks (DNNs), neuromorphic computing and non-von Neumann computing paradigms.
\end{IEEEbiography}
\vspace{+10pt}

\begin{IEEEbiography}[{\includegraphics[width=1in,height=1.25in,clip,keepaspectratio]{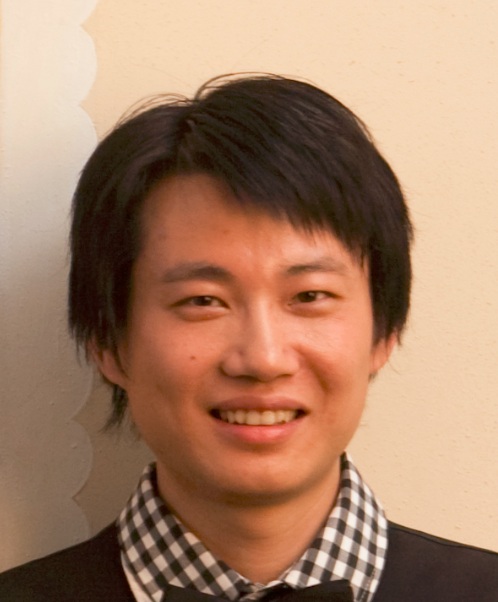}}]{Tianshi Chen}
% \begin{IEEEbiographynophoto}{Tianshi Chen}
% \todo{find recent photos}
received the bachelor's degree in mathematics from the Special Class for the Gifted Yong, University of Science and Technology of China (USTC), Hefei, China, in 2005, and the PhD degree in computer science from the Department of Computer Science and Technology, University of Science and Technology of China, Hefei, China, in 2010. He received the China Computer Federation Distinguished Doctoral Dissertation Award, in 2011 and the Chinese Academy of Sciences Distinguished Doctoral Dissertation Award, in 2011 for his PhD work on computational complexity analysis of evolutionary algorithms. He is currently a professor with the Institute of Computing Technology, Chinese Academy of Sciences, Beijing, China. He is also serving as the CEO of a startup called Cambricon Technologies Corporation Limited, whose commercial processor products are named “Cambricon”.
\end{IEEEbiography}

%%%%%%%%%% -------- end biography ----------- %%%%%%%%%%%%%%

\end{document}